\documentclass[twocolumn,letterpaper]{IEEEAerospaceCLS}  %

\usepackage{geometry}

\usepackage{graphicx} %
\usepackage{float} %
\usepackage{subcaption} %
\usepackage{wrapfig} %

\usepackage{booktabs} %
\usepackage{multirow}
\usepackage{multicol}

\usepackage{xcolor} %
\usepackage[utf8]{inputenc} %
\usepackage{amsmath} %
\usepackage{amsfonts} %
\usepackage{amssymb} %
\usepackage{amstext} %
\usepackage{mathrsfs} %
\usepackage{mathtools} %
\usepackage{accents} %
\usepackage{bbm} %
\usepackage[makeroom]{cancel} %
\usepackage{gensymb} %
\usepackage{siunitx} %
\usepackage[most]{tcolorbox} %

\usepackage[colorlinks=false,hidelinks]{hyperref} %
\usepackage{url}
\usepackage{enumitem} %

\usepackage{algorithm}
\usepackage{algpseudocode}

\newcommand{\ignore}[1]{}  %

\definecolor{customcolor}{RGB}{11,61,145}
\newcommand{\link}[2]{\textcolor{customcolor}{\href{#1}{#2}}}

\tcbuselibrary{theorems}
\newtcbtheorem[number within=section]{example}{Example}{
    colback=white,
    colframe=black,
    colbacktitle=gray!10!white,
    coltitle=black,
    fonttitle=\small\bfseries,
    fontupper=\small,
    description font=\small\bfseries,
    sharp corners,
    titlerule=0mm,
    boxrule=0.05mm,
    boxsep=3pt,
    left=2pt,
    right=2pt,
    top=2pt,
    bottom=2pt,
    before=\vspace{0.0pt},
    after=\vspace{0.1pt},
}{ex}

\newcommand{\refm}[1]{\mathcal{\uppercase{#1}}} %
\newcommand{\vecm}[4]{\prescript{\refm{#4}}{}{\mathbf{#1}}_{#2, #3}} %
\newcommand{\rotm}[3]{[\prescript{\refm{#3}}{}{\mathbf{#1}}_{\refm{#2}}]} %
\newcommand{\tram}[4]{\vecm{#1}{\refm{#2}}{\refm{#3}}{\refm{#4}}} %
\newcommand{\velm}[3]{\prescript{\refm{#3}}{}{\mathbf{#1}}_{#2}} %

\newcommand{\quatm}[3]{\prescript{\refm{#3}}{}{\mathbf{#1}}_{#2}} %
\newcommand{\quatmr}[2]{{\mathbf{#1}}_{#2}} %
\newcommand{\quatms}[2]{{\boldsymbol{#1}}_{#2}} %

\begin{document}
\title{
    Modeling Considerations for Developing Deep Space Autonomous Spacecraft and Simulators \\ \vspace{-4pt}
    {\small\bfseries Project page: \link{https://sites.google.com/stanford.edu/spacecraft-models}{sites.google.com/stanford.edu/spacecraft-models}} \vspace{-3pt}
}

\author{
    \href{https://www.chrisagia.com}{Christopher Agia}\textsuperscript{1*}, 
    \href{https://www.guillemc.com}{Guillem Casadesus Vila}\textsuperscript{2}, 
    \href{https://www-robotics.jpl.nasa.gov/who-we-are/people/saptarshi_bandyopadhyay}{Saptarshi Bandyopadhyay}\textsuperscript{3},
    \href{https://scienceandtechnology.jpl.nasa.gov/people/d_bayard}{David S. Bayard}\textsuperscript{3},\\
    \href{https://descanso.jpl.nasa.gov/biography/cheung.html}{Kar-Ming Cheung}\textsuperscript{3},
    \href{https://ieeexplore.ieee.org/author/37336718000}{Charles H. Lee}\textsuperscript{3},
    \href{https://ieeexplore.ieee.org/author/37285842700}{Eric Wood}\textsuperscript{3},
    \href{https://www.linkedin.com/in/ian-aenishanslin-a46b61175/?originalSubdomain=fr}{Ian Aenishanslin}\textsuperscript{4},
    \href{https://www.linkedin.com/in/steven-ardito-08a705b}{Steven Ardito}\textsuperscript{3},\\
    \href{https://www.cs.ucla.edu/lorraine-fesq}{Lorraine Fesq}\textsuperscript{3},
    \href{https://scholar.google.com/citations?user=RhOpyXcAAAAJ\&hl=en}{Marco Pavone}\textsuperscript{2}, 
    \href{https://www-robotics.jpl.nasa.gov/who-we-are/people/issa_nesnas}{Issa A. D. Nesnas}\textsuperscript{3}\\\\ \vspace{-2pt}
    {\small \textsuperscript{1}Department of Computer Science, Stanford University, California, U.S.A.}\\
    {\small \textsuperscript{2}Department of Aeronautics \& Astronautics, Stanford University, California, U.S.A.}\\
    {\small \textsuperscript{3}Jet Propulsion Laboratory, California Institute of Technology, California, U.S.A.}\\
    {\small \textsuperscript{4}Institut Polytechnique des Sciences Avancées, Ivry-sur-Seine, France}\\\\ \vspace{-2pt}
    {\normalsize *Corresponding author. Email: \href{mailto:cagia@stanford.edu}{\texttt{cagia@stanford.edu}}}
    \thanks{\footnotesize 979-8-3503-0462-6/24/$\$31.00$ \copyright2024 IEEE. This research was carried out at the Jet Propulsion Laboratory, California Institute of Technology, under a contract with the National Aeronautics and Space Administration. This work was also supported by the National Aeronautics and Space Administration under the Innovative Advanced Concepts (NIAC) program.}
}

\maketitle
\thispagestyle{plain}
\pagestyle{plain}

\begin{abstract}
Over the last two decades, space exploration systems have incorporated increasing levels of onboard autonomy to perform mission-critical tasks in time-sensitive scenarios or to bolster operational productivity for long-duration missions.
Such systems use \textit{models} of spacecraft subsystems and the environment to enable the execution of autonomous functions (\textit{functional-level autonomy}) within limited time windows and/or with constraints. 
These models and constraints are carefully crafted by experts on the ground and uploaded to the spacecraft via prescribed safe command sequences for the spacecraft to execute.
Such practice is limited in its efficacy for scenarios that demand greater operational flexibility.

To extend the limited scope of autonomy used in prior missions for operation in distant and complex environments, there is a need to further develop and mature autonomy that jointly reasons over multiple subsystems, which we term \textit{system-level autonomy}.
System-level autonomy establishes situational awareness that resolves conflicting information across subsystems, which may necessitate the refinement and interconnection of the underlying spacecraft and environment onboard models.
However, with a limited understanding of the assumptions and tradeoffs of modeling to arbitrary extents, designing onboard models to support system-level capabilities presents a significant challenge.
For example, simple onboard models that exclude cross-subsystem effects may compromise the efficacy of an autonomous spacecraft, while complex models that capture interdependencies among spacecraft subsystems and the environment may be infeasible to simulate under the real-world operating constraints of the spacecraft (\textit{e.g.}, limited access to spacecraft and environment states, and computational resources).

In this paper, we provide a detailed analysis of the increasing levels of model fidelity for several key spacecraft subsystems, with the goal of informing future spacecraft functional- and system-level autonomy algorithms and the physics-based simulators on which they are validated. 
We do not argue for the adoption of a particular fidelity class of models but, instead, highlight the potential tradeoffs and opportunities associated with the use of models for onboard autonomy and in physics-based simulators at various fidelity levels.
We ground our analysis in the context of deep space exploration of small bodies, an emerging frontier for autonomous spacecraft operation in space, where the choice of models employed onboard the spacecraft may determine mission success. 
We conduct our experiments in the Multi-Spacecraft Concept and Autonomy Tool (MuSCAT), a software suite for developing spacecraft autonomy algorithms.
\end{abstract} 

\tableofcontents

\section{Introduction}\label{sec:introduction} 

\begin{figure}
\centering
\includegraphics[width=3.25in]{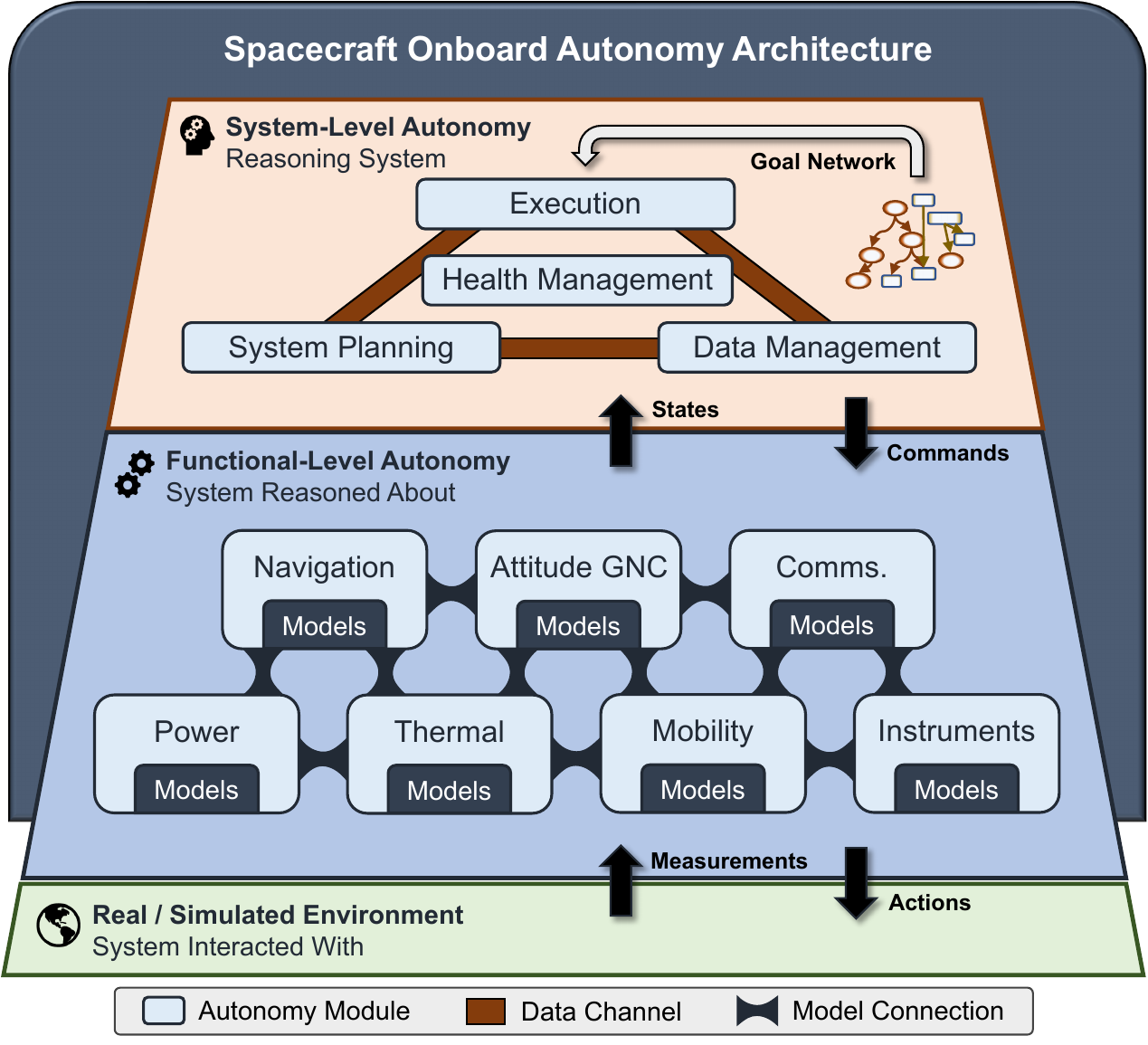}\\ 
\caption{
    \textbf{Models for spacecraft onboard autonomy.} 
    Models are a fundamental component of spacecraft onboard autonomy. 
    They describe the behavior of onboard functional-level subsystems (blue), which in turn enables the development of functional-level autonomy algorithms including planning, prediction, estimation, control. 
    Future missions to worlds with uncertain and dynamic environments will require system-level autonomy (orange) to develop and maintain situational awareness in nominal and off-nominal scenarios. 
    Such system-level capabilities must aggregate and resolve information across various functional-level subsystems and at multiple time-scales, thereby also relying on the quality of the underlying models for robust reasoning and decision making.
}
\label{fig:SC-models}
\vspace{-3pt}
\end{figure}

There is an escalating demand for spacecraft that feature increasing levels of onboard autonomy, defined as the ability of the spacecraft to achieve mission goals independent of external control (\textit{i.e.}, ground control)~\cite{fong2018autonomous}.
The need for spacecraft onboard autonomy is motivated by enabling new exploration missions, increasing productivity, enhancing robustness, and eventually reducing operations cost~\cite{nesnas2021autonomy}.
For example, the exploration of distant worlds with dynamic environments that are not well characterized \textit{a priori} may not be feasible with state-of-the-practice ground-in-the-loop operations.
In such situations, large uncertainties and communication constraints reduce the ability of ground experts to assess the states of the spacecraft and environment, predict outcomes, and prescribe command sequences in a timely manner.
Some future missions may only be viable with onboard decision making, reasoning, and taking actions that achieve goals while assuring spacecraft safety, each of which is predicated upon establishing robust situational awareness.
Other missions may benefit from increased productivity and robustness driven by onboard autonomy capable of reducing uncertainties and carrying out risk- and time-sensitive tasks across various mission phases.

Developing adequate \textit{models} of spacecraft subsystems (also referred to as \textit{subsystem models}) is among the most important technical challenges toward increasing spacecraft autonomy~\cite{nesnas2021autonomy}.
These models are essential components of both functional-level and system-level autonomy\footnote{Functional-level autonomy operates within a subsystem to produce locally autonomous behavior. It is typically implemented through the use of estimators, controllers, and local state machines. System-level autonomy operates across multiple subsystems. It develops situational awareness by aggregating information across subsystems, resolving conflicting information that may arise, assessing the health of the system, managing data, planning, scheduling, and executing tasks in both nominal and off-nominal scenarios.} (see Figure~\ref{fig:SC-models}).
They describe subsystem behavior and thereby form the basis of functional-level autonomy algorithms (\textit{e.g.}, consider how an attitude dynamics model facilitates attitude control). 
They also predict subsystem behavior, which enables system-level capabilities such as planning and system health management~\cite{kolcio2016model}.
Importantly, models that capture the interdependencies among subsystems and the environment can be used to reduce uncertainties and promote safe decision making~\cite{Ref:Feather2002quantitative}.

Nevertheless, a core outstanding challenge in using subsystem models for reliable system-level capabilities is establishing the appropriate level of \textit{model fidelity}.
Models used in previous missions (for independent functional-level autonomy tasks) may be inadequate, considering that system-level autonomy could necessitate the refinement and interconnection of subsystem models
to capture cross-subsystem dependencies and fine-grained environment effects; providing, in turn, the situational awareness required for robust reasoning and decision making.
However, arbitrarily increasing model fidelity also introduces tradeoffs.
For instance, highly accurate models (a) might rely on spacecraft and environment states that cannot be observed or accurately estimated onboard the spacecraft; (b) may be computationally intractable to run under the resource constraints of the spacecraft; (c) may complicate the design of system-level autonomy algorithms; (d) could have prediction accuracies exceeding those required for the intended system-level control goals.
These modeling considerations drive the need to categorize and analyze the different tiers of subsystem model fidelity from the perspective of system-level autonomy and its requirements.

In this paper, we outline the various degrees in which spacecraft subsystems can be modeled, with the goal of informing future subsystem modeling efforts, the development of system-level autonomy algorithms, and the design of simulators on which spacecraft onboard autonomy is validated.
\textbf{Our contributions are three-fold:} (i) a qualitative analysis of possible subsystem models that may be required onboard, which lists the progressive tiers of model fidelity and identifies where couplings exist among four major spacecraft subsystems: power, attitude GNC, navigation, and communications; (ii) a discussion of modeling trades, which serves as a basis for future trade studies on the quantitative relevance of model fidelity for system-level autonomy tasks; (iii) a case study in simulation for the cruise-approach phases of a deep-space exploration mission, demonstrating that low-fidelity models can indeed be interconnected for autonomous rendezvous with small bodies.
We conclude that considering the interconnections among subsystem models provides opportunities for cross-subsystem analysis and direct modeling---a promising pathway to enabling robust situational awareness.
Overall, this paper takes a first step toward a holistic, system-level assessment of multiple spacecraft subsystems and their associated models toward enabling greater autonomous spacecraft capabilities.

\section{Related Work}\label{sec:related-work} 

\begin{figure*}
\centering
\includegraphics[width=\textwidth]{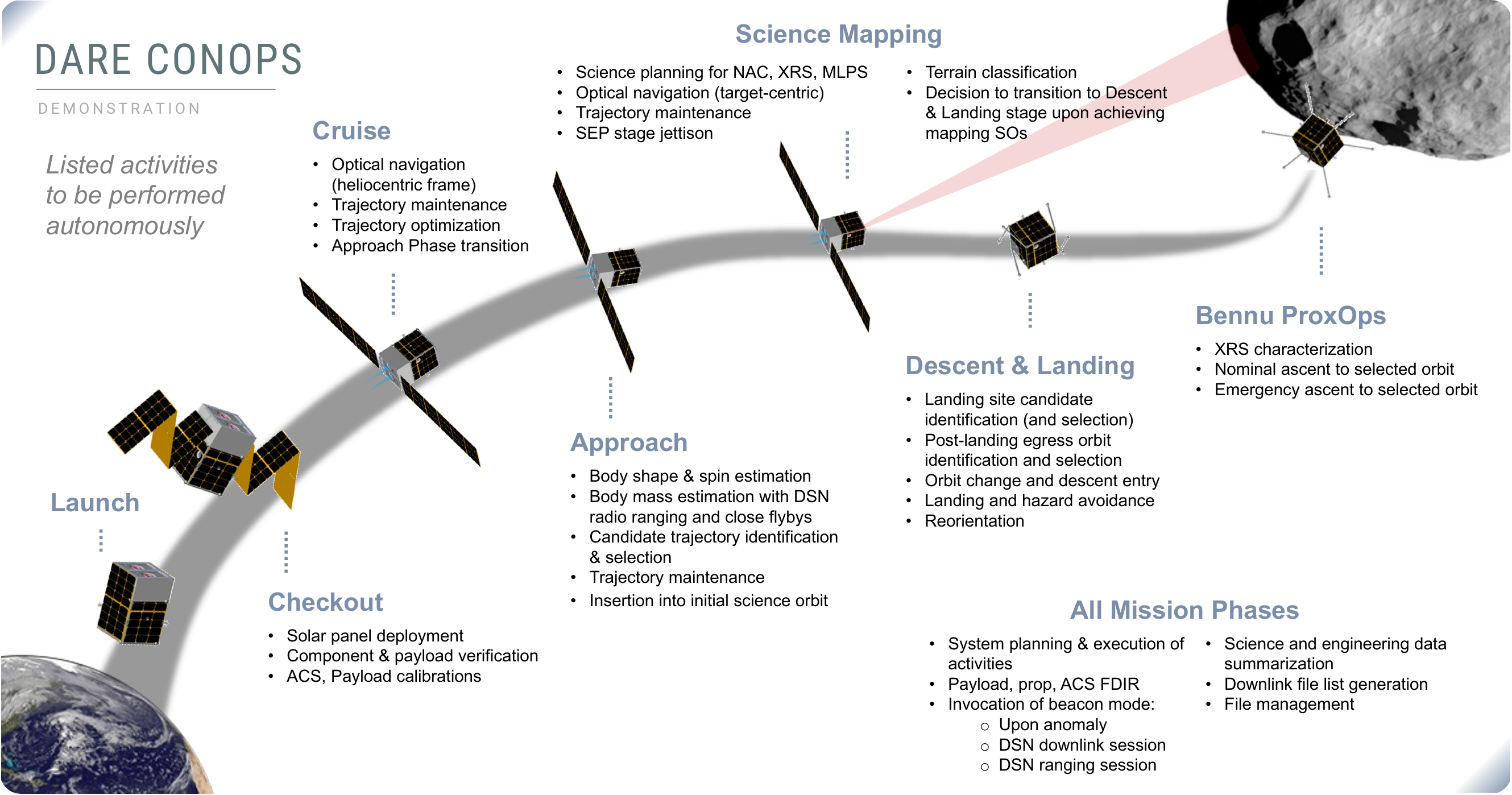}\\
\caption{
    \textbf{Autonomous small body exploration mission.} 
    We ground our analysis of subsystem model fidelity in the context of deep space exploration of small bodies with an autonomous SmallSat; a mission concept based on the Deep-space Autonomous Robotic Explorer (DARE) project at NASA JPL. 
    While DARE consists of all mission phases ranging from cruise to proximity operations, we focus our assessment on the cruise and approach mission phases. 
    These mission phases reflect nominal operating conditions suitable for assessing the role of model fidelity in four major spacecraft subsystems.
}
\label{fig:SB-mission}
\end{figure*}

\subsection{Spacecraft onboard autonomy}\label{sec:related-work-autonomy}
Spacecraft onboard functional-level autonomy has been implemented in a number of deep-space missions to carry out mission-critical tasks and increase operational productivity. 
For example, JPL's autonomous optical navigation system (AutoNav) was used for autonomous cruise and flyby tracking on the Deep Space 1 spacecraft in 1999~\cite{Ref:Riedel2000using}.
AutoNav was also used for autonomous terminal-phase navigation of the impactor and flyby spacecraft in the Deep Impact mission in 2005~\cite{Ref:Kubitschek2006deep}. 
More recently, small-body missions such as Hayabusa~2~\cite{Ref:Ogawa2020image} and OSIRIS-REx (Origins, Spectral Interpretation, Resource Identification, and Security–Regolith Explorer) \cite{Ref:Norman2022autonomous} used onboard autonomy for precise descent to small-body surfaces, while DART (Double Asteroid Redirection Test) \cite{Ref:Tropf2023dart} used onboard autonomy to guide impact with the asteroid. 
Autonomous Entry, Descent, and Landing (EDL) has been used to land the Mars Exploration Rovers (MER), Mars Science Laboratory (MSL), and Mars~2020 rovers on the Red Planet~\cite{Ref:Nelessen2019mars}.

Autonomous system-level capabilities have also been demonstrated in space. 
For example, onboard planning, scheduling, and execution was used on the Earth Orbiter (EO) 1 spacecraft~\cite{Ref:Sherwood2004operating}.
Newer algorithms for flexible execution of onboard task plans have been featured aboard ASTERIA (Arcsecond Space Telescope Enabling Research In Astrophysics) CubeSat~\cite{Ref:Fesq2021results}.
To our knowledge, no spacecraft has demonstrated end-to-end autonomy in deep space for extended portions of the mission timeline and under complex interactions with the environment.

\subsection{Modeling for spacecraft simulation}\label{sec:related-work-models}
Developing and maturing spacecraft onboard autonomy requires understanding the interconnections of models across subsystems, modeling them for novel or improved autonomous capabilities (\textit{i.e.}, situational understanding), and evaluating them in an integrated simulation platform that can test cross-subsystem interactions for accuracy and robustness.
A number of high-fidelity simulation software exists, including MONTE (Mission analysis, Operations and Navigation Toolkit Environment)~\cite{Ref:Evans2018monte} for navigation and mission design, Thermal Desktop~\cite{welch1999automating} for thermals, MMPAT (Multi-Mission Power Analysis Tool)~\cite{wood2012multi} for power, GIST (GN\&C Integrated Simulation Testbed) for guidance and control, and GNU Radio~\cite{Ref:GNURadio} for communication.
JPL's Dynamics And Real-Time Simulation (DARTS) software includes high-fidelity position and attitude dynamics while offering the capability to integrate models of other subsystems~\cite{Ref:DARTS}.
However, these simulators are either domain specific or do not natively integrate models across multiple subsystems.

Low-fidelity simulation platforms such as Modelon~\cite{Ref:Modelon}, Omniverse~\cite{Ref:Omniverse}, STK~\cite{Ref:STK}, Gazebo~\cite{Ref:GAZEBO}, and Basilisk~\cite{Ref:Basilisk} are suitable for simulating some spacecraft subsystems, but they do not represent a multi-subsystem spacecraft simulator and most have not been validated with real data.
To our knowledge, an integrated simulation architecture combining models across multiple core spacecraft subsystems does not exist. 
Many spacecraft undergo high-fidelity integrated modeling of their subsystems when the spacecraft is being built (\textit{e.g.}, the structural thermal optical performance (STOP) modeling of the James Webb Space Telescope (JWST)~\cite{Ref:Mcelwain2023james}), but these models are not used in the autonomy development phase.

\section{Case Study: Autonomous Exploration of Small Bodies}\label{sec:case-study}
We aim to outline the various levels of onboard model fidelity for several key spacecraft subsystems.
We conduct our analysis in the context of a deep space exploration mission, which grounds our results and discussion in a realistic autonomy use case.
Specifically, we consider an autonomous rendezvous mission of a SmallSat with a distant small body based on the \textbf{D}eep-space \textbf{A}utonomous \textbf{R}obotic \textbf{E}xplorer (DARE) project at NASA JPL.
A visualization is provided in Figure~\ref{fig:SB-mission}.

\subsection{Selected mission rationale: Small body cruise-approach}\label{sec:mission}
Small-body exploration missions have been proposed as a promising avenue to mature spacecraft autonomy algorithms and architectures at relatively low costs \cite{nesnas2021autonomy}. 
Due to their large numbers (nearly a million to date), small sizes, and vast distance ranges from Earth's telescopes, the ephemerides, rotational parameters, and physical properties of small bodies are often not accurately known. 
Moreover, their low mass and irregular shapes induce weak, non-uniform gravity fields, producing complex but low-magnitude disturbances within their field of influence.
Small-body missions thus contain several sources of uncertainty that challenge current state-of-the-art spacecraft autonomy while offering opportunities for incremental improvement.  

While small-body exploration missions could span multiple mission phases, including cruise, approach, proximity operations, landing, and surface operations, in this paper, we focus exclusively on the cruise and approach phases.
During these two mission phases, the spacecraft onboard autonomy must (a) estimate the spacecraft's orbit, (b) execute trajectory correction maneuvers (TCMs) to maintain course on the pre-planned mission trajectory toward the small body, (c) estimate and control the spacecraft attitude to ensure proper solar charging, communication with Earth, and de-saturation of reaction wheels, and (d) estimate and control other spacecraft states to manage thermal and data subsystems. 
The selected mission phases encompass coupling among various spacecraft subsystems.
As such, they are suitable for a first-pass analysis of subsystem model fidelity and autonomous spacecraft operation in nominal conditions.

\subsection{Experiment details: Spacecraft and onboard autonomy}\label{sec:experiments}
We briefly describe the spacecraft design, subsystems, and onboard autonomy algorithms considered in the case study.

\subsubsection{Spacecraft design and components}
\begin{figure}
\centering
\includegraphics[width=3.25in]{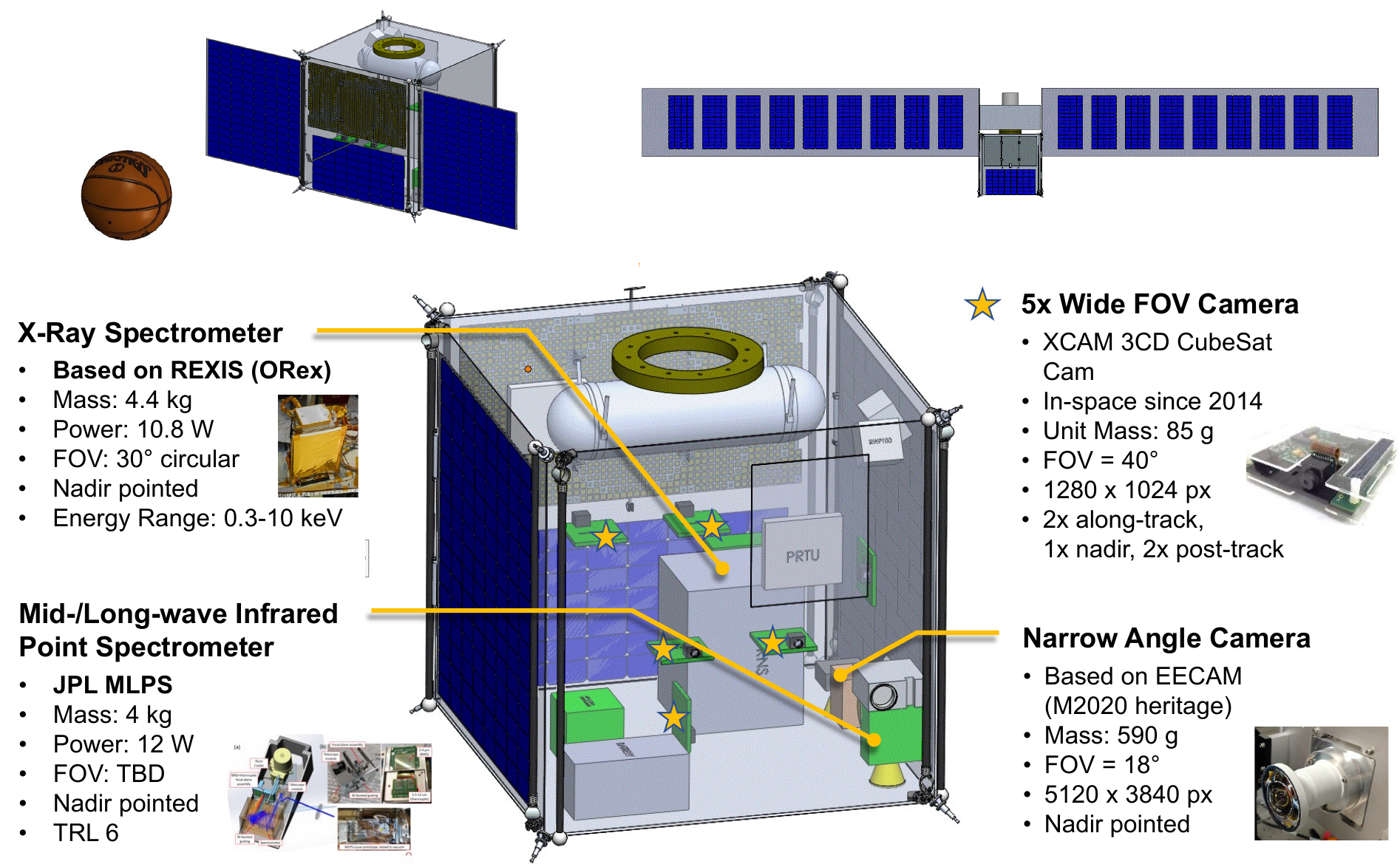}\\
\caption{
    \textbf{Spacecraft design for small body exploration.} 
    For the autonomous rendezvous mission with a small body, we design a SmallSat spacecraft that consists of the main subsystems we analyze in this work. 
    The spacecraft is powered by five solar arrays and equipped with several actuators for linear and angular control including a solar electric propulsion thruster, microthrusters, and reaction wheel assemblies.
}
\label{fig:SC-smallsat}
\end{figure}

For the small-body mission, we leverage the design of a hypothetical SmallSat spacecraft from the DARE project.
The $\SI{30}{\cm}^3$ spacecraft consists of two separable units: (a) a solar-electric propulsion (SEP) stage consisting of a gimbaled thruster for linear motion and two large articulated solar arrays, and (b) the main stage equipped with four reaction-wheel assemblies and two micro thrusters for fully-actuated angular control.
The spacecraft is powered by a total of five solar arrays: three small body-mounted arrays and the two large solar wings attached to the SEP stage. 
Power generated by the solar arrays is transferred directly to active electrical components, and excess energy is stored in two $\SI{40}{Wh}$ lithium-ion batteries. 
The spacecraft's maximum expected weight is $\SI{178}{\kg}$ (wet mass), putting it in the range of the Minisatellite class of SmallSats~\cite{papais2020architecture}.
The spacecraft is visualized in Figure~\ref{fig:SC-smallsat}.

For communications, the spacecraft is equipped with a dipole antenna operating at $\SI{8450}{\mega\hertz}$ with a $\SI{28.1}{\decibel}$ gain and a beam width of $\SI{0.1}{\degree}$.
It has a transmission power of $\SI{50}{\watt}$, a line loss of $\SI{1}{\decibel}$, and a noise temperature of $\SI{100}{\kelvin}$.
We assume a coding gain of $\SI{7.3}{\decibel}$, an energy per bit of $\SI{4.2}{\decibel}$, and a data-rate limit of $\SI{1}{\mega\byte\per\second}$.

The perception sensors and instrument payload include five wide-angle cameras and one narrow-angle camera, a mid- and long-wave infrared point spectrometer~\cite{chen2022mid}, and an x-ray spectrometer~\cite{hong2021calibration}.
Only the sensors and instruments that are pointed toward the small body are \textit{active} (\textit{i.e.}, powered on).
Because instrument models are not considered in this work, we do not generate science observations but instead simulate the power and memory consumption of instruments during the mission.
Science and engineering data is stored in a $\SI{1}{\giga\byte}$ onboard memory drive.

\subsubsection{Onboard subsystems}
We study the power, attitude GNC, navigation, and communications subsystems operating under nominal cruise and approach conditions to a distant small body.
For each subsystem, we discuss low-fidelity implementations of spacecraft onboard models, considerations for high-fidelity modeling, and the implications of modeling on spacecraft onboard autonomy.

Since science mapping during the proximity phase, descent and landing, and surface operations are beyond the scope of this work, we do not address the instrument and surface mobility subsystems.
Depending on the operating conditions of the spacecraft, the thermal subsystem may have significant coupling and effect on the other subsystems, which we will address in follow-on publications.

\begin{figure*}
\centering
\includegraphics[width=0.85\textwidth]{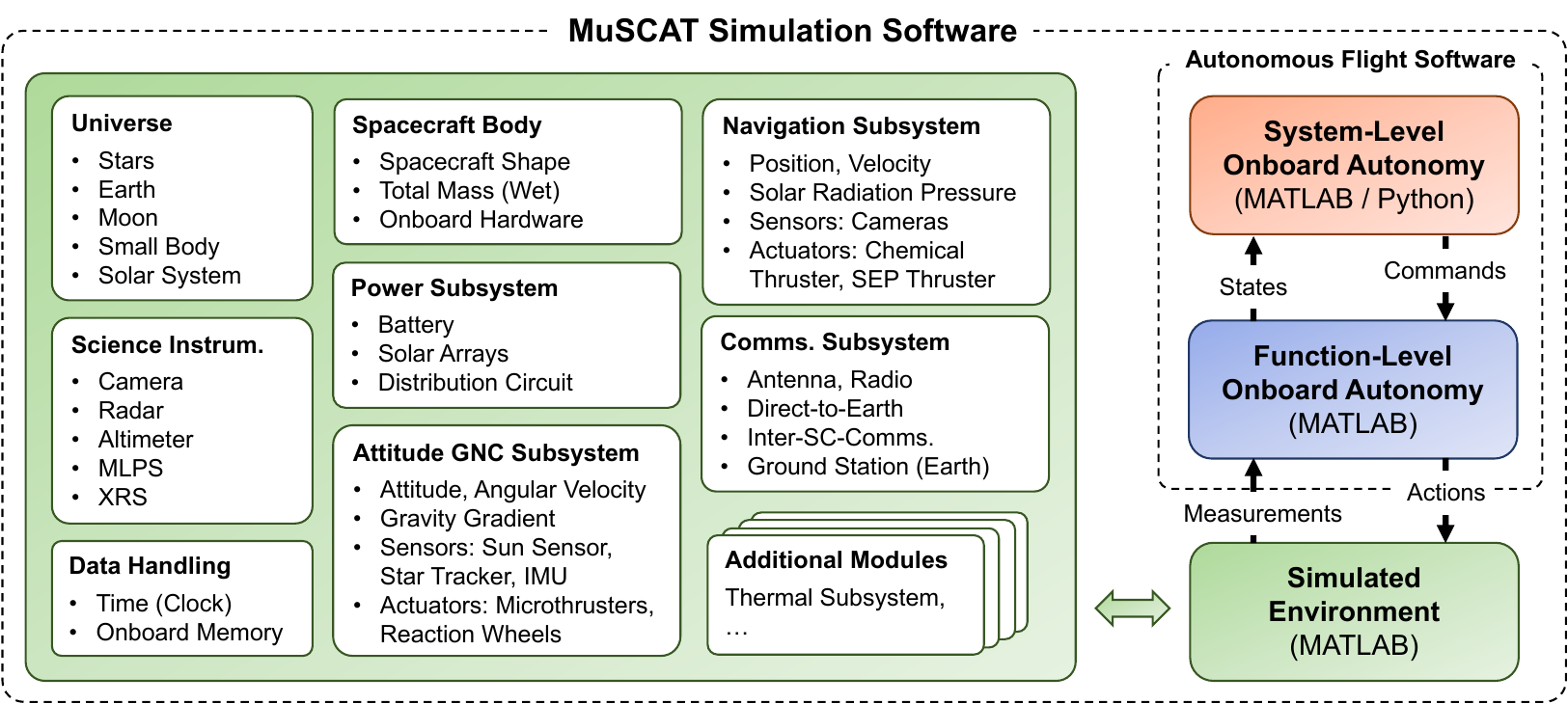}\\
\caption{\textbf{MuSCAT simulation software.} We run our experiments in MuSCAT, a simulation tool that implements low-fidelity spacecraft and environment models, autonomous onboard flight software (both functional-level and system-level capabilities), science instruments and payload, and other spacecraft onboard components. Different from existing simulators, which often target high-fidelity simulation of a single spacecraft subsystem, MuSCAT integrates low-fidelity models across multiple spacecraft subsystems to support the prototyping and simulation of mission concepts that may benefit from onboard autonomy.}
\label{fig:muscat}
\end{figure*}

\subsubsection{Onboard functional-level autonomy}
Each onboard subsystem contains functional-level autonomy algorithms for handling subsystem state(s): estimating current/past state(s), controlling state, predicting future state(s), and planning future actions to constrain future state(s).
In the simplest case, a subsystem may directly use low-fidelity models for estimation. For instance, in the power subsystem, we can estimate the battery state-of-charge (SoC) at low fidelity by accumulating net power margins at each timestep.
Within the attitude GNC subsystem, we use attitude guidance based on the eigen-axis method~\cite{wertz2012spacecraft} for three-axis reorientation, where quaternion multiplication is used to compute the unique rotation axis and angle for the minimum path maneuver.
Tracking is performed using a nonlinear attitude controller that converges exponentially (in the presence of modeling uncertainties and disturbances) to attitude states computed by attitude guidance~\cite{bandyopadhyay2016nonlinear}.
For Navigation, TCMs are computed using a Lambert orbital transfer maneuver~\cite{lancaster1969unified,izzo2015revisiting}.

Attitude GNC and navigation estimators are not considered in our experiments.
Instead, we assume ground-truth access to the linear and angular states of the spacecraft and the small body.
Nonetheless, we discuss models used for estimation in their respective subsystem sections below.

\subsubsection{Onboard system-level autonomy}
At the system-level, the \textit{executive} is the top-level orchestrator of the spacecraft. 
It schedules and dispatches tasks based on their priority to achieve goals sent from the ground and/or elaborated by the onboard planner (as illustrated in Figure~\ref{fig:SC-models}). 
The executive also monitors the state of the spacecraft and assesses its progress toward mission objectives and goals. Examples of tasks include setting attitude modes for charging or pointing to the small body, issuing TCMs, or establishing ground communication for science and engineering data downlink. 

For our prototype, we implemented a system-level executive that maintains a priority queue of the presented autonomy tasks. 
We consider a simple event-triggered scheduling of activities.
For example, if during operation, the battery SoC drops below a specified minimum threshold (\textit{i.e.}, the event), the executive will schedule a charging cycle, whereby a heuristic approach is used to compute a new spacecraft attitude, balancing power generated by the solar arrays and the pointing error to the small body.
For simplicity, the priorities of each activity are hand-specified for our mission.
Recharging is the highest priority activity, as the spacecraft needs power to remain viable, followed by executing TCMs and, finally, performing communication downlink.

We note that more sophisticated system-level autonomy algorithms for activity planning and execution~\cite{troesch2020mexec}, health management~\cite{kolcio2016model,djebko2019model}, and data management~\cite{rabideau2014data} may be required for more complex missions.
For example, autonomously executing all phases of the DARE mission, including science mapping, descent, landing, and proximity operations (as shown in Figure~\ref{fig:SB-mission}), will require an expanded set of system-level autonomy capabilities.
Our event-triggered software executive is sufficient for studying the behavior of the subsystem models during the small body cruise and approach.

\subsection{Simulation software: MuSCAT}\label{sec:muscat}
We run experiments in the Multi-Spacecraft Concept and Autonomy Tool (MuSCAT)\footnote{The MuSCAT simulator is designed to support multi-spacecraft missions but can just as well be used for single spacecraft missions. This work exclusively considers its use for a single spacecraft small-body mission.}. 
MuSCAT is an internal simulation software developed at NASA JPL.
It implements \textit{low-fidelity} models for simulation across the four subsystems we analyze in this paper and is thus an efficient test bed for mission concepts such as our small-body case study.
A visualization of MuSCAT is provided in Figure~\ref{fig:muscat}.

Low-fidelity models implemented in MuSCAT relate to subsystem models that may run onboard the spacecraft.
For example, MuSCAT's low-fidelity attitude dynamics are identical to the onboard model from which our attitude controller is derived~\cite{bandyopadhyay2016nonlinear}. 
However, it is not typically the case that spacecraft simulation software and onboard subsystems employ identical models, as their modeling considerations differ.
Section~\ref{sec:discussion} discusses the differences in modeling considerations for spacecraft simulators versus onboard subsystems.

\section{Preliminaries}
We provide a brief overview of the mathematical conventions used hereafter for the expression of subsystem models and onboard autonomy algorithms.

Reference frames or \textit{frames} are represented with caligraphic font.
For example, we use $\refm{i}$ to represent the J2000 inertial frame centered at Earth's origin.
Other commonly used frames include the spacecraft frame $\refm{sc}$, Sun frame $\refm{s}$, and small-body frame $\refm{b}$.
We express transformations between frames with a rotation and translation.
For example, the position and orientation of the spacecraft with respect to J2000 (read ``from spacecraft to J2000'') is defined by rotation matrix $\rotm{R}{sc}{i}$ in $\mathrm{SO}(3)$ and translation vector $\tram{r}{i}{sc}{i}$ in ${\mathbb{R}}^3$.
Finally, we denote time derivatives with the dot notation; the spacecraft's velocity with respect to J2000 is given by $\velm{\dot{r}}{\refm{sc}}{i}$.

The following sections present multiple possible models for describing the behavior of a subsystem at different levels of fidelity.
To distinguish fidelity levels, we use hat notation to denote quantities associated with low-fidelity models.
For instance, the state output of a low-fidelity model may be expressed as $\hat{\mathbf{x}}$, while its high-fidelity counterpart will be $\mathbf{x}$.

\section{Power Subsystem}\label{sec:power}
The power subsystem is responsible for generating, distributing, and storing power onboard the spacecraft.
Modeling the power subsystem enables tracking, control, and prediction of the spacecraft's power profile given current and future states and actions.
It involves modeling photovoltaic power generation, power distribution and losses, and battery SoC prediction, among others.
Such capabilities are beneficial for both functional- and system-level autonomy.
For example, functional-level autonomy operating within the power subsystem may regulate the current input to the battery for safe charging.
Likewise, for system-level autonomy, the scheduler may rely on battery SoC predictions to determine how long the spacecraft's solar arrays should be oriented toward the sun before enough energy is accumulated for a planned TCM.

The following sections will discuss modeling trades associated with different processes in the power subsystem.
For a discussion summary, please refer to Table~\ref{tab:power-subsystem}.

\subsection{Power generation: Solar arrays}\label{sec:power-generation}
In our case study, solar arrays are the sole producer of energy onboard the spacecraft.
Let us first consider an example of a low-fidelity onboard model for the power generated by $J$ solar arrays.
Each solar array has an area $a_j$, a solar cell efficiency $e_j$, and a packing fraction $p_j$.
The centroid of each solar array is located at $\vecm{r}{{\refm{s}}}{j}{i} = \tram{r}{s}{sc}{i} + \rotm{R}{sc}{i}\vecm{r}{{\refm{sc}}}{j}{sc}$ with a surface normal unit vector $\prescript{\refm{sc}}{}{\mathbf{\hat{n}}}_j$.
We can compute the angle of incidence of the sunlight on the $j$-th solar array as:
\begin{equation}\label{eq:power-solar-angle-lf}
    \theta_j = \arccos\left(\frac{\vecm{r}{{\refm{s}}}{j}{i}}{||\vecm{r}{{\refm{s}}}{j}{i}||^2} \cdot \rotm{R}{sc}{i} \prescript{\refm{sc}}{}{\mathbf{\hat{n}}}_j \right).
\end{equation}
The power generated by all $J$ solar arrays is thus given by:
\begin{equation}\label{eq:power-solar-lf}
    \hat{P}_{\mathrm{solar}} = \sum_{j=1}^J H\left(||\tram{r}{s}{sc}{i}||\right) a_j e_j p_j \cos(\theta_j) \quad (\si{\watt}),
\end{equation}
where $H\left(||\tram{r}{s}{sc}{i}||\right)$ in $\si{\watt\per\meter\squared}$ is the power density of the sunlight (Eq.~\ref{eq:light-irradiance}) at the spacecraft's current distance from the sun $||\tram{r}{s}{sc}{i}||$ (further details provided in the Appendix).

\begin{figure}
\centering
\includegraphics[width=3.25in]{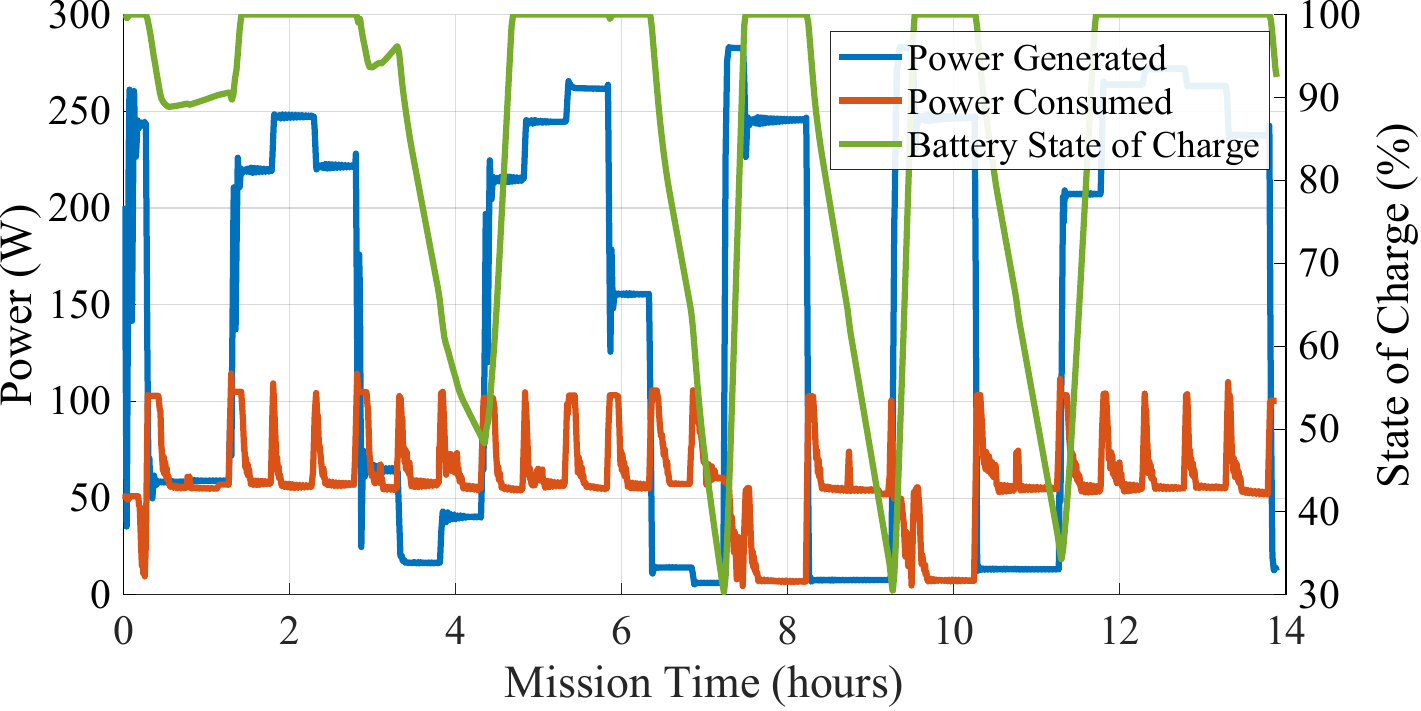}\\
\caption{
    \textbf{Power subsystem case study.} 
    Power onboard the spacecraft is generated by solar arrays and directly distributed to electrical components and batteries.
    The results above are the predictions of low-fidelity power generation, distribution, and battery state of charge models used during the small-body mission (see Figure~\ref{fig:SB-mission}).
    Generated power predictions fluctuate with the changing attitude of the spacecraft, while peaks in consumed power correspond to firing solar electric propulsion thrusters. 
    Notice that when more power is consumed than generated, the predicted battery state of charge gradually decreases until thirty percent, at which point the system-level software executive schedules an immediate charging cycle.
}
\label{fig:power}
\end{figure}

We classify this solar array model as \textit{low-fidelity} because, despite accounting for the time-varying position and attitude of the spacecraft, it does not capture more complex effects like occlusion~\cite{katzan1991effects,8997803} and lifetime degradation~\cite{wertz2011space}.
We can thus only use this model to coarsely estimate generated power, as shown in Figure~\ref{fig:power}, where we observe significant power variances throughout the small-body mission.

Let us now consider a higher-fidelity model of a solar array.
We express the voltage of the solar array in functional form:
\begin{equation}\label{eq:power-solar-voltage-mf}
    \hat{V}_{\mathrm{solar}} = f_{\hat{V}}\left(H, \theta, \xi, O, L_d, \eta\right) \quad (\si{\volt}),
\end{equation}
where $H$ is the power density as a function of the spacecraft's position, $\theta$ is the angle of incidence of the sunlight as a function of the spacecraft's attitude (Eq.~\ref{eq:power-solar-angle-lf}), $\xi$ is the efficiency of the solar array as a function of its temperature, $O$ is the shading or occlusion on the solar arrays as a function of the environment and spacecraft geometry, $L_d$ is the lifetime degradation of the solar arrays as a function of time, and $\eta$ is a constant that accounts for the solar array design parameters and assembly losses (\textit{e.g.}, area, packing fraction). 
The power generated by the solar array is then expressed as the product of its output voltage (Eq.~\ref{eq:power-solar-voltage-mf}) and the load current $I_{\mathrm{load}}$:
\begin{equation}\label{eq:power-solar-mf}
    P_{\mathrm{solar}} = \hat{V}_{\mathrm{solar}} I_{\mathrm{load}} \quad (\si{\watt}).
\end{equation}

\begin{example}{Interconnecting models for greater accuracy}{power-solar-array}
    Notice that the higher-fidelity model (Eq.~\ref{eq:power-solar-mf}) captures several important effects on solar arrays that were neglected in the low-fidelity model (Eq.~\ref{eq:power-solar-lf}). 
    The key insight is that refining models to capture high-fidelity effects drives the need to interconnect models across subsystems.
    Consider how solar array efficiency $\xi$ depends on temperature readings from the thermal subsystem.
\end{example}

\begin{table}
\renewcommand{\arraystretch}{1.4}
\caption{
    \textbf{Power subsystem modeling trade.} 
    An overview of the low- and high-fidelity modeling considerations for three different processes in the power subsystem. 
    For each listed model parameter, items to the right of the colon are dependencies. 
    For example, the low-fidelity solar array efficiency ($e$) is constant, while its high-fidelity counterpart ($\xi$) depends on spacecraft temperature. 
    Mathematical symbols in parenthesis correspond to quantities introduced in the main text.
}
\label{tab:power-subsystem}
\centering
\scriptsize
    \begin{tabular}{cll}
    \hline
    \textbf{} & \textbf{Low fidelity power} & \textbf{High fidelity power} \\
    \hline
    \hline
    \multirow{5}{*}{\rotatebox[origin=c]{90}{Solar array}} 
        & Incidence angle ($\theta$): S/C att. & Occlusion ($O$): S/C \& Env. \\
        & Power density ($H$): S/C pos. & MPPT: I-V curves, \ldots \\ 
        & Area ($a$): const. & Lifetime degrade ($L_d$): time \\
        & Packing fraction ($p$): const. & Design \& losses ($\eta$): const. \\
        & Efficiency ($e$): const. & Efficiency ($\xi$): S/C temp. \\
    \hline
    \multirow{2}{*}{\rotatebox[origin=c]{90}{Usage}} 
        & Net power ($\hat{P}_{\mathrm{net}}$): time & Bus conv. loss ($\mu$): temp., \ldots \\
        & Power rating ($P_c$): const. &  - \\
    \hline
    \multirow{4}{*}{\rotatebox[origin=c]{90}{Battery}} 
        & Net Energy ($\hat{E}_{\mathrm{net}}$): time & Coulomb count ($\int I_b$): time \\
        & Max capacity ($\hat{E}_{\mathrm{max}}$): const. & Max capacity ($E_{\mathrm{max}}$): time \\
        & Charge eff. ($e_b^+$): const. & - \\
        & Discharge eff. ($e_b^-$): const. & - \\
    \hline
    \multicolumn{3}{l}{\tiny\bfseries S/C: Spacecraft, MPPT: Maximum Power Point Tracking, Env.: Environment} \\[-1.0ex]
    \multicolumn{3}{l}{\tiny\bfseries att.: attitude, pos.: position, const.: constant, eff.: efficiency, temp.: temperature, conv.: conversion}
    \end{tabular}
\end{table}

We can build on Eq.~\ref{eq:power-solar-voltage-mf} to construct a model of the non-linear I-V Characteristics (IVCs) of the solar array.
Due to the intrinsic properties of solar arrays, their output voltage drops as the load (current) increases, resulting in less generated power~\cite{wertz2011space}.
This IVC coupling can be captured by modeling output voltage and as a function of load current:
\begin{equation}\label{eq:power-solar-voltage-hf}
    V_{\mathrm{solar}} = f_{V}\left(H, \theta, \xi, O, L_d, \eta, I_{\mathrm{load}}\right) \quad (\si{\volt}).
\end{equation}
Eq.~\ref{eq:power-solar-voltage-hf} shows that the IVCs of the solar array also shift based on the variables outlined Eq.~\ref{eq:power-solar-voltage-mf}.
Several approaches can be used to implement this model onboard the spacecraft.
In this simplest case, a non-parametric approach can be taken by storing empirical (tabular) data of the IVCs in onboard memory.
For a new set of variables (\textit{e.g.}, occlusions, temperature), the data can be interpolated for the output voltage. 
More recent works take analytic~\cite{zhang2017prediction} and learning-based~\cite{khatib2018new} approaches to modeling IVCs, which are promising avenues for predicting generated power onboard the spacecraft.

As a final consideration, we can attempt to capture the effects of maximum power point tracking (MPPT).
MPPT regulates the load current on the solar arrays to avoid voltage drops and maximize generated power (Eq.~\ref{eq:power-solar-mf}).
It is considered a form of onboard functional-level autonomy operating within the power subsystem.
By monitoring the output voltage of the solar array and adjusting the load current accordingly, MPPT can track the shifting IVCs of the solar array in real time.

\begin{example}{Modeling tradeoffs for onboard autonomy}{power-mppt}
    Modeling IVCs (Eq.~\ref{eq:power-solar-voltage-hf}) of the solar array captures the coupling between output voltage and load current (in contrast to Eq.~\ref{eq:power-solar-mf}), which in turn enables MPPT to maximize generated power via onboard control algorithms.
    This example highlights that understanding the fidelity tradeoffs associated with different models can yield opportunities to develop or improve onboard autonomy.
\end{example}

While the effects of MPPT are significant, yielding between ten and forty percent gains in generated power~\cite{wertz2011space}, modeling the MPPT process can be challenging. 
An analytic model will depend on the specific implementation of the power subsystem, including the solar arrays, the MPPT control law, and the circuit design.
Given an IVC model (Eq.~\ref{eq:power-solar-voltage-hf}), one simple option is to cast MPPT as the optimization problem:
\begin{equation}\label{eq:power-solar-mppt}
    P_{\mathrm{solar}}^* = \max_{I_{\mathrm{load}}} \left(V_{\mathrm{solar}} I_{\mathrm{load}}\right) \quad (\si{\watt}).
\end{equation}
In practice, this model will produce \textit{optimistic} predictions of generated power for several reasons.
Real-world imperfections in the MPPT hardware, including noise, component tolerances, and transient response times, can prevent precise tracking of the maximum power point, thus causing deviations from the idealized model.

\subsection{Power distribution: Usage and losses}\label{sec:power-distribution}
The role of power distribution is to route power, ensuring a reliable supply to all payloads and electrical components onboard the spacecraft, and to manage power by switching components off when appropriate to conserve energy.

We first consider an example of a low-fidelity onboard model for power distribution that quantifies the net instantaneous power in the spacecraft.
At each discrete time step $t$, power is generated by the solar arrays and consumed by all \textit{active} spacecraft components.
Let $C$ be the number of power-consuming spacecraft components (\textit{e.g.}, sensors, instruments, actuators).
The net instantaneous power is expressed as:
\begin{equation}\label{eq:power-net-lf}
    \hat{P}_{\mathrm{net}} = P_{\mathrm{solar}} - \sum_{c=1}^C {\mathbbm{1}}_{c} P_{c} \quad (\si{\watt}),
\end{equation}
where ${\mathbbm{1}}_c(t)$ is an indicator function that equals $1$ if component $c$ is active at time $t$ (otherwise equals $0$) and $P_c$ is the constant power rating of component $c$.
Treating components as constant power loads is valid because their power ratings typically integrate voltage conversion losses.
This power distribution model provides coarse estimates of consumed power onboard the spacecraft.
The corresponding results from our small-body mission are shown in Figure~\ref{fig:power}.

We classify the power distribution model in Eq.~\ref{eq:power-net-lf} as low-fidelity because its accuracy decreases with the use of MPPT.
Concretely, solar power generated by MPPT undergoes voltage conversion losses~\cite{meade2008parasitic} before reaching a suitable bus voltage for battery charging and powering components.
Ignoring these losses can result in overly optimistic predictions of net power.
A high-fidelity model must capture the bus voltage conversion loss on generated power before distribution:
\begin{equation}\label{eq:power-net-hf}
    P_{\mathrm{net}} = \mu P^*_{\mathrm{solar}} - \sum_{c=1}^C {\mathbbm{1}}_{c} P_{c} \quad (\si{\watt}).
\end{equation}
Here, $P^*_{\mathrm{solar}}$ is the generated power from MPPT (Eq.~\ref{eq:power-solar-mppt}).
The bus voltage conversion loss, denoted by $\mu$, is influenced by factors such as load current, input-output voltage, and operating temperature.
The specific voltage converter's efficiency curves typically illustrate variations based on these factors.

Finally, given the net power of the spacecraft (Eq.~\ref{eq:power-net-hf}), the energy surplus (or deficit) over a $\delta t$ time step is:
\begin{equation}\label{eq:power-net-energy-lf}
    \hat{E}_{\mathrm{net}}(t) = \frac{1}{3600} \int_{t - \delta t}^{t} P_{\mathrm{net}}(\tau) d\tau \quad (\si{Wh}).
\end{equation}

\subsection{Power storage: Battery charge/discharge \& SoC}\label{sec:power-storage}
Onboard models associated with power storage are used to estimate battery charge rates, discharge rates, and SoC. 
For the small-body mission, we specifically use lithium-ion batteries, while other battery types are also possible.

Let us first consider a low-fidelity model for battery charge and discharge.
We assume a constant charging efficiency $e^+_b$, discharging efficiency $e^-_b$, and maximum battery capacity $\hat{E}_{\mathrm{max}}$.
The change in battery SoC $\Delta \hat{Q}_b(t)$ can be modeled as:
\begin{equation}\label{eq:power-charge-lf}
    \Delta \hat{Q}_b(t) = \begin{cases}
        e_b^+ \hat{E}_{\mathrm{net}}(t) / \hat{E}_{\mathrm{max}}, \ \text{if} \ \hat{E}_{\mathrm{net}}(t) > 0 \\[5pt]
        \hat{E}_{\mathrm{net}}(t)/(e_b^- \hat{E}_{\mathrm{max}}), \ \text{if} \ \hat{E}_{\mathrm{net}}(t) \le 0
    \end{cases},
\end{equation}
where $\hat{E}_{\mathrm{net}}$ is the energy surplus or deficit of the spacecraft (Eq.~\ref{eq:power-net-energy-lf}) at time step $t$.
Figure~\ref{fig:power} shows the battery SoC estimates from this model over the small-body mission duration.

This battery model is considered low-fidelity because it assumes linear charge characteristics and a fixed maximum battery capacity.
These assumptions simplify modeling at the cost of potentially large deviations from the true battery SoC.
In reality, batteries have (a) non-linear charge characteristics and (b) a varying maximum capacity, both of which shift with usage, thermal cycling, and operating temperatures.
A high-fidelity battery model must account for both (a) and (b) to provide accurate estimates of battery SoC.

Coulomb Counting techniques~\cite{murnane2017closer,movassagh2021critical} directly integrate current flow in and out of the battery, which captures non-linear charge characteristics. 
These techniques are implemented using dedicated hardware (\textit{i.e.}, current integrators), which makes them suitable for onboard battery SoC estimation but does not enable model-based prediction. 
We express the change in battery SoC with Coulomb Counting as:
\begin{equation}\label{eq:power-charge-hf}
    \Delta Q_b(t) = \frac{\int^t_{t-\delta t} I_b(\tau)d\tau}{3600 \ E_{\mathrm{max}}(t)},
\end{equation}
where $I_b$ is the current flowing in (or out) of the battery and $E_{\mathrm{max}}$ is its time-varying maximum capacity.
In theory, $E_{\mathrm{max}}$ can be estimated with knowledge of the battery cycle life and an accurate degradation model.
As such, both analytic and data-driven models are being explored for battery cycle life~\cite{ning2004cycle,su2021cycle,tarar2023accurate} and degradation~\cite{de2023mathematical,lu2022battery}.

\begin{example}{Model tradeoffs for onboard autonomy}{power-battery}   
    While various models can be used to express a subsystem process, their autonomy use-case depends on the implementation of the model. 
    Consider how Coulomb Counting (Eq.~\ref{eq:power-charge-hf}) is implemented with dedicated hardware, which may limit its autonomy use to onboard estimation tasks. 
    In contrast, the simpler model in Eq.~\ref{eq:power-charge-lf} may be more amenable to onboard prediction tasks since it only relies on predictions of generated and consumed power.
\end{example}

Finally, estimation of battery SoC onboard the spacecraft can be strengthened by combining Coulomb Counting with direct battery SoC prediction methods, for which both analytic~\cite{parhizi2020analytical} learning-based~\cite{chemali2018state,ng2020predicting} models are being developed.

\section{Attitude GNC Subsystem}\label{sec:attitude}
The attitude GNC subsystem is responsible for managing and altering the attitude state of the spacecraft via estimation, planning, and control algorithms.
It includes attitude kinematics and dynamics models, which describe how the attitude state of the spacecraft evolves with respect to disturbance and control torques.
Variations of these models are designed to capture a range of effects on the spacecraft's attitude.
Typically, low-fidelity attitude dynamics form the basis of onboard functional-level autonomy algorithms in the attitude GNC subsystem, while higher-fidelity models are used for verification and validation of autonomy in simulation.

The following sections will discuss modeling trades associated with different processes in the attitude GNC subsystem.
For a discussion summary, please refer to Table~\ref{tab:attitude-subsystem}.

\subsection{Low-fidelity attitude modeling: Rigid-body dynamics}\label{sec:attitude-lf}
We begin with a standard rigid-body attitude dynamics model with $n$ reaction wheels (or \textit{rotors}) and $m$ gimballed solar wings. 
This takes the form of a non-linear dynamical system:
\begin{equation}\label{eq:att-dynamical-system}
    \dot{\mathbf{x}}^{\mathrm{att}} = {\mathbf{f}}^{\mathrm{att}}\left({\mathbf{x}}^{\mathrm{att}}, {\mathbf{u}}^{\mathrm{att}}\right) \quad \text{where} \quad {\mathbf{x}}^{\mathrm{att}} = \begin{bmatrix}
        \quatm{q}{sc}{i} \\
        \quatms{\omega}{sc} \\
        \quatms{\omega}{rw} \\
        \quatms{\omega}{sw}
     \end{bmatrix},
\end{equation}
where $\mathbf{x}^{\mathrm{att}}$ in ${\mathbb{R}}^{7+n+m}$ is the attitude state of the spacecraft, which consists of its quaternion attitude $\quatm{q}{sc}{i}$ in ${\mathbb{R}}^4$ and angular velocity $\quatms{\omega}{sc}$ in ${\mathbb{R}}^3$ expressed in the inertial reference frame, the scalar angular velocities of all $n$ reaction wheels $\quatms{\omega}{rw}$ in ${\mathbb{R}}^n$, and the scalar angular velocities of all $m$ gimballed solar wings $\quatms{\omega}{sw}$ in ${\mathbb{R}}^m$.
The control input $\mathbf{u}^{\mathrm{att}}$ in ${\mathbb{R}}^3$ corresponds to the total torque applied by the attitude actuators.
For example, our small-body mission employs twelve microthrusters and four reaction wheels.

We express the full dynamical system (Eq.~\ref{eq:att-dynamical-system}) in terms of its state derivatives $\dot{\mathbf{x}}^{\mathrm{att}}$. 
The first term of the state derivative is the quaternion rate of the spacecraft $\quatmr{\dot{q}}{sc}$, which is obtained through the following attitude kinematic model:
\begin{equation}\label{eq:att-kinematics-lf}
    \quatmr{\dot{q}}{sc} = \frac{1}{2} {\boldsymbol{\Omega}}\left(\quatms{\omega}{sc}\right) \quatm{q}{sc}{i} = \frac{1}{2} {\boldsymbol{\Xi}}\left(\quatm{q}{sc}{i}\right) \quatms{\omega}{sc},
\end{equation}
where, for any given quaternion ${\mathbf{q}}$ composed of its vector and scalar components as $[{\mathbf{q}}_v, q_s]^T$, we have:
\begin{equation*}
    {\boldsymbol{\Omega}}\left(\boldsymbol{\omega}\right) \triangleq \begin{bmatrix}
        -{\boldsymbol{\omega}}^\times & {\boldsymbol{\omega}} \\
        -{\boldsymbol{\omega}}^T & 0
    \end{bmatrix}\,; \;  
    {\boldsymbol{\Xi}}\left(\mathbf{q}\right) \triangleq \begin{bmatrix}
        q_s {\mathbf{1}}_{3\times3} + {\mathbf{q}}_v^\times \\
        -{\mathbf{q}}_v^T
    \end{bmatrix}.
\end{equation*}
Here, ${\boldsymbol{x}}^\times$ denotes the cross-product matrix defined as:
\begin{equation*}
    {\boldsymbol{x}}^\times \triangleq \begin{bmatrix}
        0 & -x_3 & x_2 \\
        x_3 & 0 & -x_1 \\
        -x_2 & x_1 & 0
    \end{bmatrix}
\end{equation*}

\begin{table}
\renewcommand{\arraystretch}{1.4}
\caption{
    \textbf{Attitude GNC subsystem modeling trade.} 
    An overview of the low-fidelity and high-fidelity modeling considerations for the attitude GNC subsystem. 
    The most salient tradeoff of low-fidelity onboard models is that they neglect flexible body dynamics. 
    For example, deflection of solar wings and vibrations from attitude actuators. 
    Robustness to these effects is often \textit{built-in} to onboard functional-level autonomy for attitude guidance, estimation, and control.
}
\label{tab:attitude-subsystem}
\centering
\scriptsize
    \begin{tabular}{cll}
    \hline
    \textbf{} & \textbf{Low fidelity attitude} & \textbf{High fidelity attitude} \\
    \hline
    \hline
    \multirow{6}{*}{\rotatebox[origin=c]{90}{Dynamics}} 
        & Rigid body dynamics & Flexible body dynamics \\
        & Reaction wheels ($\quatms{\omega}{rw}$): time & Reaction wheel vibration \\
        & Solar wings ($\quatms{\omega}{sw}$): time & Solar wing flex \& vibration \\ 
        & Gravity dist. (${\mathbf{T}}^g$): S/C pos., att. & Microthruster vibration \\
        & SRP dist. (${\mathbf{T}}^\rho$): S/C pos., att. & - \\
        & S/C Inertia (${\mathbf{I}}_{cm}$): wet mass & - \\
    \hline
    \multirow{4}{*}{\rotatebox[origin=c]{90}{Integration}} 
        & Quaternion normalization & - \\
        & Quaternion properization & - \\ 
        & Const. axis of rot. sampling rate & - \\ 
        & Const. quaternion rate sampling rate & - \\ 
    \hline
    \multicolumn{3}{l}{\tiny\bfseries S/C: Spacecraft, SRP: Solar Radiation Pressure, Const.: Constant} \\[-1.0ex]
    \multicolumn{3}{l}{\tiny\bfseries att.: attitude, pos.: position, rot.: rotation, dist.: disturbance}
    \end{tabular}
\end{table}

\begin{figure}
\centering
\includegraphics[width=3.25in]{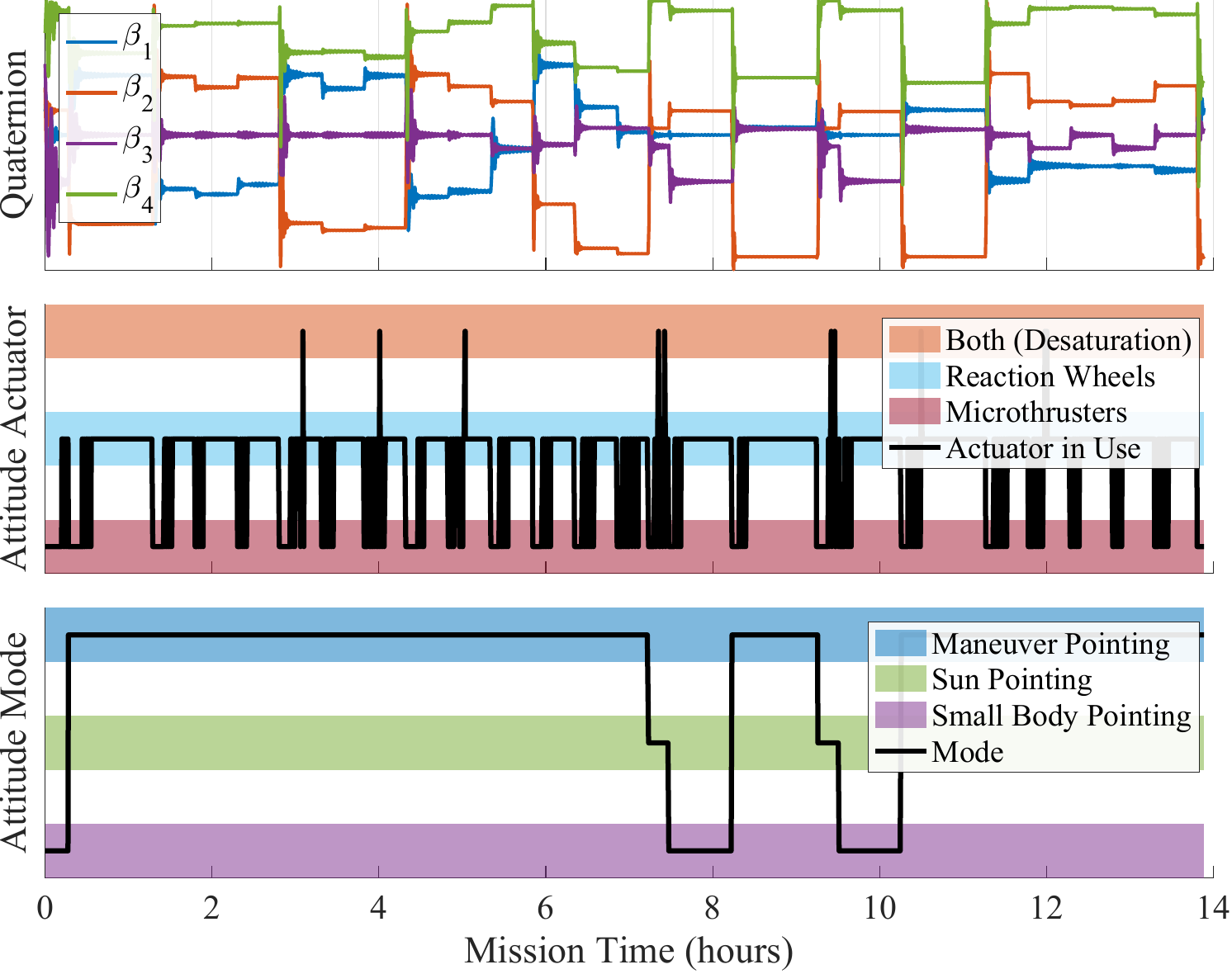}\\
\caption{
    \textbf{Attitude GNC subsystem case study.}
    The attitude state of the spacecraft is managed by a system-level software executive and altered by functional-level attitude GNC algorithms (see architecture in Figure~\ref{fig:SC-models}).
    The results show the performance of autonomous guidance and control algorithms developed on low-fidelity models of attitude dynamics.
    The first plot shows stable tracking control of attitude states using a combination of actuators shown in the second plot.
    The last plot shows attitude modes set by the system-level software executive to achieve mission goals and ensure system health.
}
\label{fig:attitude}
\end{figure}

The second term of the state derivative $\dot{\mathbf{x}}^{\mathrm{att}}$ (Eq.~\ref{eq:att-dynamical-system}) is the angular acceleration of the spacecraft $\quatms{\dot{\omega}}{sc}$, which is obtained through the following rigid-body attitude dynamics model:
\begin{equation}\label{eq:att-dynamics-lf}
    \quatms{\dot{\omega}}{sc} = {\mathbf{I}}_{cm}^{-1} \left(\prescript{\refm{sc}}{}{\mathbf{T}}^d + {\mathbf{u}}^{\mathrm{att}} - \quatms{\omega}{sc}^\times \left({\mathbf{J}}_{cm}\quatms{\omega}{sc} + {\mathbf{h}}_{\omega}\right)\right),
\end{equation}
where ${\mathbf{I}}_{cm}$ in ${\mathbb{R}}^{3\times3}$ is the inertia matrix of the spacecraft about its center of mass, ${\mathbf{J}}_{cm}$ in ${\mathbb{R}}^{3\times3}$ is the inertia matrix including the inertia contribution of the reaction wheels and gimballed solar wings, and $\prescript{\refm{sc}}{}{\mathbf{T}}^d$ in ${\mathbb{R}}^3$ is the disturbance torque on the system.  
${\mathbf{h}}_{\omega}$ in ${\mathbb{R}}^3$ is the angular momentum vector of the reaction wheels and solar wings, expressed as:
\begin{equation*}
    {\mathbf{h}}_{\omega} = \sum_{i=1}^n {\mathbf{I}}_{rw, i} {\mathbf{a}}_{rw, i} \quatms{\omega}{rw, i}  + \sum_{j=1}^m {\mathbf{I}}_{sw, j} {\mathbf{a}}_{sw, j} \quatms{\omega}{sw, j},
\end{equation*}
where ${\mathbf{I}}_{rw, i}, {\mathbf{I}}_{sw, i}$ in ${\mathbb{R}}^{3\times3}$ are the inertia matrices, ${\mathbf{a}}_{rw, i}, {\mathbf{a}}_{sw, i}$ in ${{\mathbb{R}}^3}$ are axis of rotation (unit) vectors, and 
$\quatms{\omega}{rw, i}, \quatms{\omega}{sw, j}$ in ${\mathbb{R}}$ are the angular velocities of the $n$ reaction wheels and $m$ solar wings, respectively.
Lastly, we consider the \textit{torque form} interpretation of ${\mathbf{J}}_{cm}$ and $\mathbf{u}^{\mathrm{att}}$ given in~\cite{bayard2010high}:
\begin{equation*}
    {\mathbf{J}}_{cm} = {\mathbf{I}}_{cm} - \sum_{i=1}^n {\mathbf{I}}_{rw, i} - \sum_{j=1}^m {\mathbf{I}}_{sw, j},
\end{equation*}
\begin{equation}\label{eq:att-control-input}
    \begin{split}
        {\mathbf{u}}^{\mathrm{att}} &= -\sum_{i=1}^n {\mathbf{I}}_{rw, i} \left({\mathbf{a}}_{rw, i} \quatms{\dot{\omega}}{rw, i} + \quatms{\dot{\omega}}{sc} \right) \\[-8pt]
        &\quad\quad\quad\quad - \sum_{j=1}^m {\mathbf{I}}_{sw, j}\left({\mathbf{a}}_{sw, j} \quatms{\dot{\omega}}{sw, j} + \quatms{\dot{\omega}}{sc} \right),
    \end{split}
\end{equation}
where $\quatms{\dot{\omega}}{rw, i}, \quatms{\dot{\omega}}{sw, j}$ in ${\mathbb{R}}$ are the angular accelerations of the reaction wheels and solar wings, respectively.

The rigid-body attitude dynamics model given by Eq.~\ref{eq:att-dynamics-lf} is commonly used in the design and analysis of onboard functional-level autonomy in the attitude GNC subsystem. 
For example, it is used to derive attitude controllers~\cite{bayard2010high,bandyopadhyay2016nonlinear}, whereby control torques ${\mathbf{u}}^{\mathrm{att}}$ are computed to track a desired attitude state with desirable guarantees (\textit{e.g.}, stability) and dynamical responses subject to disturbance torques $\prescript{\refm{sc}}{}{\mathbf{T}}^d$.
Eq.~\ref{eq:att-control-input} can then be solved via optimization to obtain the reaction wheel accelerations required for the maneuver.
Similar techniques are used for microthruster-based actuation and reaction wheel desaturation procedures.
For large maneuvers, attitude guidance algorithms~\cite{5281102,ponche2023guidance} profile the turn by decomposing the desired final attitude state into a series of states that the attitude controller tracks sequentially.

Figure~\ref{fig:attitude} exemplifies attitude guidance and control techniques operating in the context of our small-body mission.
The mission starts at approximately $\SI{300}{\kilo\meter}$ distance from the small body.
Hence, we consider two disturbance sources: the gravitational disturbance torque $\prescript{\refm{sc}}{}{\mathbf{T}}^g$ induced by the Sun and small body (Eq.~\ref{eq:gravity-net-torque-approx}), and the solar radiation pressure (SRP) disturbance torque $\prescript{\refm{sc}}{}{\mathbf{T}}^\rho$ induced by the Sun (Eq.~\ref{eq:srp-net-torque}).
We observe that attitude modes issued by the system-level software executive, such as recharging, small-body pointing, and TCM modes, are successfully tracked using a combination of microthrusters and reaction wheels for actuation.

There are also cases where the rigid-body dynamics model (Eq.~\ref{eq:att-dynamics-lf}) is not appropriate for the design of onboard autonomy algorithms.
For example, it is typically not used for non-linear attitude estimation algorithms because its prediction accuracy diminishes with the relatively large noise in control torques~\cite{crassidis2007survey,yang2017attitude} generated by microthrusters (\textit{e.g.}, Cassini had approximately ten percent thrust noise).
Instead, filtering-based attitude estimators often directly integrate gyroscope measurements for the attitude \textit{propagation} step~\cite{crassidis2016three}.

Beyond onboard autonomy, the full dynamical system Eq.~\ref{eq:att-dynamical-system} can also be used for rigid-body simulation. 
For instance, the low-fidelity simulator we use for our experiments, MuSCAT, employs an equivalent model for attitude simulation.
Given a time step size $\delta t$, numerical integration is performed to predict the next attitude state of the spacecraft:
\begin{equation}\label{eq:att-integrate-lf}
    {\mathbf{x}}_{t + \delta t}^{\mathrm{att}} = \int_{t}^{t+\delta t} {\mathbf{f}}^{\mathrm{att}}({\mathbf{x}}^{\mathrm{att}}_\tau, {\mathbf{u}}^{\mathrm{att}}_\tau) d\tau + {\mathbf{x}}_t^{\mathrm{att}},
\end{equation}
noting that with any significant additive operation on the state ${\mathbf{x}}^{\mathrm{att}}$, the quaternion attitude $\quatm{q}{sc}{i}$ must be normalized such that $||\quatm{q}{sc}{i}||$ sums to one and properized to maintain a positive scalar component $q_s$.
For higher accuracy, the sampling rate used by the ODE solver should be set such that the quaternion rate (Eq.~\ref{eq:att-kinematics-lf}) or the axis of rotation is approximately constant over the integration window.

\subsection{Considerations for high-fidelity attitude modeling}\label{sec:attitude-hf}
In our experiments, the attitude GNC models used in MuSCAT are equiavalent to the models used to derive the non-linear attitude controller in \cite{bandyopadhyay2016nonlinear}.
As such, we directly inherit the performance guarantees of the attitude controller proved under the rigid-body dynamics model (\textit{i.e.}, global exponential convergence).
However, the presence of complex real-world effects not captured by rigid-body dynamics can significantly impact the performance of onboard functional-level autonomy algorithms.
For example, actuator vibration and large slew maneuvers can induce small and large angle deflection of the spacecraft's solar wings~\cite{herber2014reducing}.
If unaccounted for, these complex effects can extend post-maneuver settling times or destabilize the attitude controller~\cite{deyst1969survey,wie1984attitude}.
The state-of-the-practice preemptively accounts for these effects in the control design phase, where loop shaping techniques are used to achieve desired frequency response characteristics according to performance and robustness specifications (\textit{i.e.}, ensuring sufficient gain, phase, and flex margins).

\begin{example}{Modeling tradeoffs for onboard autonomy}{attitude-control}
    Where low-fidelity models offer convenience in the initial design of onboard autonomy algorithms, it is common practice to \textit{build-in} robustness to real-world effects not captured by low-fidelity models.
    Consider how attitude controllers are carefully tuned to account for flexible body effects despite being derived from rigid-body dynamics.
    In this case, recognizing the limitations of the low-fidelity model and understanding what high-fidelity effects must be accounted for are key to robust autonomy design.
\end{example}

It is also possible to directly construct flexible body dynamics models for spacecraft attitude using finite element methods (FEMs)~\cite{wei2017dynamic}.
We consider such models high-fidelity due to their accuracy, which comes at the cost of high computational complexity, rendering them unsuitable for spacecraft onboard autonomy.
Instead, flexible body dynamics models are suited to high-fidelity simulation, enabling the verification and validation of onboard autonomy algorithms designed for robustness to complex real-world effects (see Example~\ref{ex:attitude-control}).

Recent work focuses on connecting models across the attitude GNC and power subsystems. 
For example, photovoltaic models are used for sun pointing~\cite{colagrossi2022spacecraft} and attitude determination in rover~\cite{ishida2014attitude} and satellite~\cite{porras2022use} applications.
In particular, \cite{porras2022use} studies the performance of attitude determination algorithms under photovoltaic models of increasing fidelity, including models that capture the effects of panel temperature (as discussed in Section~\ref{sec:power-generation}).
These works demonstrate that interconnecting models can enable new onboard autonomy capabilities as well as promote greater estimation accuracy.

\begin{example}{Interconnecting models for greater awareness}{attitude-solar-estimation}
    Modeling cross-subsystem effects presents more than just an opportunity to improve the performance of existing functional-level autonomy algorithms.
    Redundant information from adjacent subsystems promotes situational awareness for system-level autonomy tasks.
    Consider how solar array voltage readings can be cross-checked against attitude estimates to detect faults.
\end{example}

Fully autonomous spacecraft operation requires that system-level autonomy understands the behavior of the underlying functional-level autonomy algorithms in nominal and off-nominal conditions.
For example, system-level autonomy must ensure that TCMs do not violate the flex margins of the solar wings or that a contingency plan does not compromise the stability of the attitude controller in an attempt to recover from a fault.
Such may not require running high-fidelity attitude GNC models onboard the spacecraft, but may instead benefit from redundancy offered by low-fidelity models.
For example, despite Eq.~\ref{eq:att-dynamics-lf} being unsuitable for attitude estimation (due to microthruster noise), it can still be used for onboard fault detection or momentum management~\cite{yang2017spacecraft} tasks. 
Overall, identifying the appropriate model fidelity for system-level autonomy tasks is an important and unsolved problem.
We leave the study of this problem to future work.

\section{Navigation Subsystem}\label{sec:navigation}
The navigation subsystem is responsible for estimating the navigation state (\textit{i.e.}, position and velocity) of the spacecraft, its relative orbit with the small body, and planning and executing TCMs for autonomous rendezvous.
It employs orbit propagation models to predict how the navigation state changes under disturbance and control forces.
Variations of these models are used to propagate orbits with higher accuracy and numerical precision, whereas simplified models are adopted for onboard autonomy.
The navigation subsystem also includes measurement and instrument models, which facilitates orbit determination, the process of estimating spacecraft and small body states through measurements, observations, and radiometric communication with ground stations.

The following sections will discuss modeling trades associated with different processes in the navigation subsystem.
For a discussion summary, please refer to Table~\ref{tab:navigation-subsystem}.

\subsection{Orbit propagation modeling}
Orbit propagation models predict how the position and velocity of the spacecraft evolve when perturbed by disturbance and propulsion forces.
These models take the form of the following non-linear dynamical system:
\begin{equation}\label{eq:nav-lf}
    \dot{\mathbf{x}}^{\mathrm{nav}} = {\mathbf{f}}^{\mathrm{nav}}\left({\mathbf{x}}^{\mathrm{nav}}, {\mathbf{u}}^{\mathrm{nav}}\right) \quad \text{where} \quad \mathbf{x}^{\mathrm{nav}} = \begin{bmatrix}
        \tram{r}{s}{sc}{i}\\
        \velm{\dot{r}}{\refm{sc}}{i}
     \end{bmatrix},
\end{equation}
where $\mathbf{x}^{\mathrm{nav}}$ in ${\mathbb{R}}^{6}$ is the navigation state, which consists of the position $\tram{r}{s}{sc}{i}$ in ${\mathbb{R}}^{3}$ and velocity $\velm{\dot{r}}{\refm{sc}}{i}$ in ${\mathbb{R}}^{3}$ of the spacecraft with respect to frame $\refm{i}$.
The control input ${\mathbf{u}}^{\mathrm{nav}}$ is the total thrust applied by onboard propulsion systems.

We express the dynamical system (Eq.~\ref{eq:nav-lf}) in terms of its state derivatives, which capture the gravitational forces induced by $K$ celestial bodies on the spacecraft, along with SRP, disturbance, and propulsion forces.
The acceleration of the spacecraft $\velm{\ddot{r}}{\refm{sc}}{i}$ is given by the following dynamics model:
\begin{equation}\label{eq:nav-dynamics-lf}
    \velm{\ddot{r}}{\refm{sc}}{i} = \frac{1}{m_{sc}} \left(\sum_{k=1}^K \vecm{F}{k}{{\refm{cm}}}{i}^g + \prescript{\refm{i}}{}{\mathbf{F}^\rho} + \prescript{\refm{i}}{}{{\mathbf{F}}^d} + \mathbf{u}^{\mathrm{nav}}\right),
\end{equation}
where $m_{sc}$ is the mass of the spacecraft, $\vecm{F}{k}{{\refm{cm}}}{i}^g$ is the gravitational force induced by the $k$-th celestial body, $\prescript{\refm{i}}{}{\mathbf{F}^\rho}$ is the SRP force, $\mathbf{u}^{\mathrm{nav}}$ is the thrust, and $\prescript{\refm{i}}{}{{\mathbf{F}}^d}$ accounts for all other disturbance forces (\textit{e.g.}, atmospheric and electromagnetic drag). 
We assume the spacecraft frame $\refm{sc}$ is located at the center of mass $\refm{cm}$, which drifts as fuel is consumed.

The overall fidelity of the orbit propagation model (Eq.~\ref{eq:nav-dynamics-lf}) is determined by what perturbation forces it accounts for and how accurately each of the perturbations are modeled.
Low-fidelity orbit propagation models could, for example, employ low-order gravity disturbance models and cannonball SRP models, where the spacecraft's shape is approximated by a sphere~\cite{farres2018high}.
Such models could be reasonably accurate when the spacecraft is operating sufficiently far from the Sun and other gravitationally interacting bodies, as in our small-body mission.
However, close-proximity navigation with small bodies requires the use of high-fidelity orbit propagation models.
For example, high-fidelity modeling of gravitational fields induced by small bodies includes higher-order spherical gravity harmonics~\cite{herrera2013modeling} and finite element methods~\cite{park2010estimating} (also used to estimate the internal density variations of small bodies).
Likewise, $N$-plate models~\cite{JEAN2019167} and, in special cases, ray tracing techniques~\cite{ziebart2001analytical} are used for high-fidelity SRP modeling, accounting for absorption and reflection effects on the spacecraft's geometry (\textit{i.e.}, given by a CAD model).

\begin{table}
\renewcommand{\arraystretch}{1.4}
\caption{
    \textbf{Navigation subsystem modeling trade.} 
    Low-fidelity models for orbit propagation are subject to inaccuracies in multi-body or close-proximity mission scenarios.
    High-fidelity considerations include further modeling the interconnections among navigation, communication, and attitude GNC, the physical properties of the spacecraft and the small body, and onboard time for greater accuracy.
}
\label{tab:navigation-subsystem}
\centering
\scriptsize
    \begin{tabular}{cll}
    \hline
    \textbf{} & \textbf{Low fidelity navigation} & \textbf{High fidelity navigation} \\
    \hline
    \hline
    \multirow{6}{*}{\rotatebox[origin=c]{90}{Orbit propagation}} 
        & Single or two-body dynamics                  & Multi-body dynamics\\
        & Low-order gravity dist. (${\mathbf{F}}^g$)   & SGH or FEM gravity dist. (${\mathbf{F}}^g$)\\ 
        & Cannonball SRP dist. (${\mathbf{F}}^\rho$)   & N-plate or RT SRP dist. (${\mathbf{F}}^\rho$)\\
        & -                                            & Atmospheric drag dist.\\
        & -                                            & Magnetic drag dist.\\
        & -                                            & Relativistic effects\\
    \hline
    \multicolumn{3}{l}{\tiny\bfseries SRP: Solar Radiation Pressure, SGH: Spherical Gravity Harmonics, FEM: Finite Element Method} \\[-1.0ex]
    \multicolumn{3}{l}{\tiny\bfseries RT: Ray Tracing, dist.: disturbance}
    \end{tabular}
\end{table}

Low-fidelity orbit propagation models are commonly used for functional-level onboard autonomy in the navigation subsystem.
For example, orbit determination algorithms based on Extended Kalman Filters (EKF)~\cite{julier1997new} approximate Eq.~\ref{eq:nav-dynamics-lf} by exclusively modeling gravity perturbations (\textit{i.e.}, assuming SRP, drag, and other perturbing forces as noise)~\cite{keil2014kalman}.
This simplifies the computation of the state transition matrix ${\mathbf{F}}^{\mathrm{nav}}$ as required by EKFs to propagate the state covariance:
\begin{gather*}
    {\mathbf{F}}^{\mathrm{nav}} = \frac{\partial \velm{\ddot{r}}{\refm{sc}}{i}}{\partial \tram{r}{s}{sc}{i}} = \sum_{k=1}^K \frac{\partial \velm{\ddot{r}}{\refm{sc}}{i}}{\partial \vecm{r}{k}{\refm{sc}}{i}} \frac{\partial \vecm{r}{k}{\refm{sc}}{i}}{\partial \tram{r}{s}{sc}{i}} \quad \mathrm{where} \\
    \frac{\partial \velm{\ddot{r}}{\refm{sc}}{i}}{\partial \vecm{r}{k}{\refm{sc}}{i}} = \frac{3\mu_k \vecm{r}{k}{\refm{sc}}{i} (\vecm{r}{k}{{\refm{sc}}}{i})^T}{||\vecm{r}{k}{\refm{sc}}{i}||_2^5} -\frac{\mu_k {\mathbf{1}}_{3\times 3}}{||\vecm{r}{k}{\refm{sc}}{i}||_2^3}.
\end{gather*}
Here, $\vecm{r}{k}{\refm{sc}}{i}$ in ${{\mathbb{R}}^3}$ is the position of the spacecraft with respect to $k$-th celestial body (equal to $\tram{r}{s}{sc}{i} - \vecm{r}{{\refm{s}}}{k}{i}$).

\begin{figure}
\centering
\includegraphics[width=3.25in]{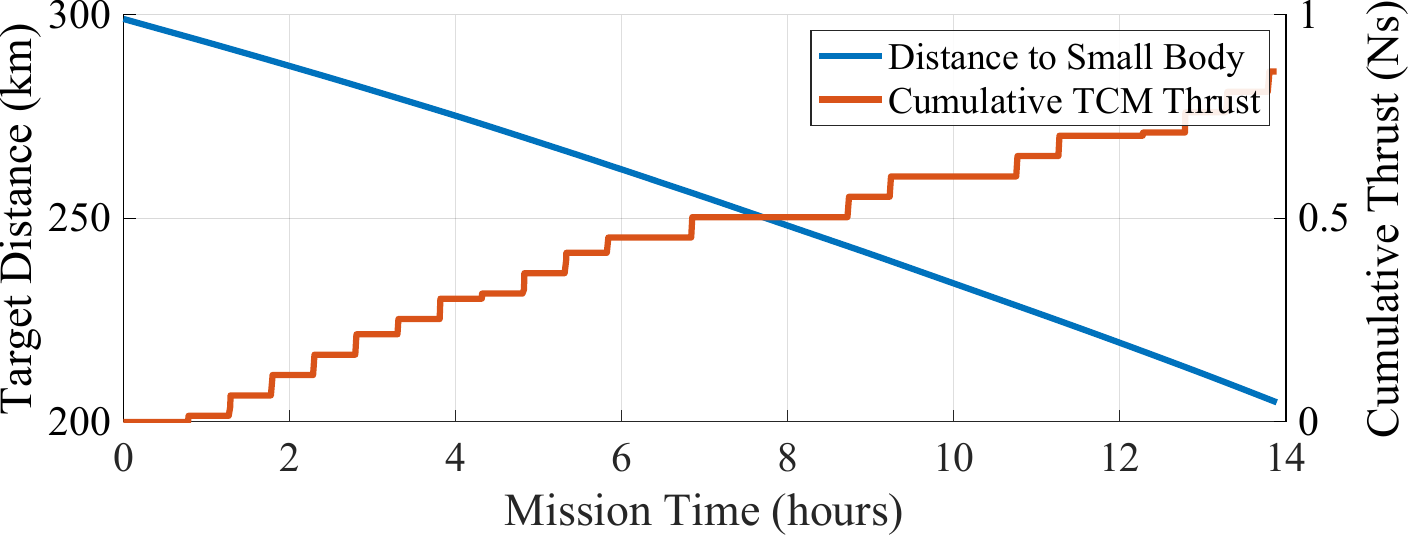}\\
\caption{
    \textbf{Navigation subsystem case study.}
    The spacecraft autonomously performs a series of trajectory correction maneuvers (TCMs) to rendezvous with the target small body. 
    These TCMs are computed using Lambert's problem, an astrodynamics model that determines the trajectory between two points within a specified time frame. 
    The result shows the spacecraft's approach with cumulative thrust from TCMs.
}
\label{fig:navigation}
\end{figure}

Low-fidelity orbit propagation models can also be used in simulation to obtain ground-truth position and velocity of the spacecraft.
For instance, MuSCAT considers (a) gravitational forces on the spacecraft induced by the Sun and the small body (Eq.~\ref{eq:gravity-force-far}), and (b) a spacecraft $N$-plate model for SRP (Eq.~\ref{eq:srp-net-force}).
Given a time step size $\delta t$, numerical integration is performed to predict the next navigation state as follows:
\begin{equation}\label{eq:nav-integrate-lf}
    {\mathbf{x}}_{t + \delta t}^{\mathrm{nav}} = \int_{t}^{t+\delta t} {\mathbf{f}}^{\mathrm{nav}}({\mathbf{x}}^{\mathrm{nav}}_\tau, {\mathbf{u}}^{\mathrm{nav}}_\tau) d\tau + {\mathbf{x}}_t^{\mathrm{nav}}.
\end{equation}
Notice that Eq.~\ref{eq:nav-integrate-lf} integrates the dynamical system in Eq.~\ref{eq:nav-lf}, where the position of the spacecraft $\tram{r}{s}{sc}{i}$ is expressed at some distance from the Sun (denoted by $\refm{s}$).
The Sun is therefore the \textit{center of integration}.
To maintain numerically precise states during simulation, the center of integration must be switched once the spacecraft enters the sphere of influence of a new celestial body, such as the small body.

The result of simulating our small-body mission with MuSCAT's orbit propagation model is shown in Figure~\ref{fig:navigation}. 
We observe that TCMs autonomously executed by the spacecraft drive it closer to the small body as desired.
This is achieved by providing the onboard autonomy with simulated spacecraft and small body states resembling those that can be estimated with optical navigation methods once the small body is within view (\textit{i.e.}, during the approach phase)~\cite{nesnas2021autonomous}.

In complex orbital scenarios, however, low-fidelity orbit propagation models can lead to cumulative numerical inaccuracies over time.
Consider future missions to Europa or Enceladus.
These missions may need to account for many gravitationally interacting bodies, atmospheric drag~\cite{nwankwo2021atmospheric}, magnetic fields~\cite{SHUVALOV202041}, and other relativistic effects.
Moreover, when dominant forces greatly overshadow minor perturbations, as seen in interactions with massive celestial bodies like gas giants, the numerical integration of low-fidelity models in Eq.~\ref{eq:nav-integrate-lf} can be susceptible to truncation errors~\cite{stiefel1969stabilization}.
Thus, increasingly complex multi-body scenarios may require the adoption of high-fidelity orbit propagation models.

\subsection{Instrument and measurement models for orbit determination}
In addition to orbit propagation models, the navigation subsystem also employs various instrument and measurement models.
These models are used by navigation or \textit{orbit determination} algorithms to estimate states of the spacecraft and the small body using onboard observations, measurements, or radiometric communication with the ground station.
Several navigation methods and models have been explored.
For example, radiometric navigation involves measuring the time delay and Doppler shift of radio signals between the spacecraft and ground station~\cite{thornton2003radiometric}, while optical navigation uses images of the target small body captured by the spacecraft~\cite{nesnas2021autonomous,Ref:Owen2011_OpNav} (autonomous optical navigation has been experimentally applied on the Deep Space 1 spacecraft~\cite{Ref:Bhaskaran1998orbit}).
Pulsar-based navigation is a novel method using timing signals from X-ray pulsars for spacecraft localization~\cite{Ref:Chen2017pulsar,xue2019x,malgarini2023application}. 
In our study using the MuSCAT simulator, rather than implementing high-fidelity instrument and measurement models for radiometric or optical navigation, we approximate their performance using realistic bounds on error and associated uncertainty in states used by other onboard algorithms.

\subsection{Considerations for cross-subsystem modeling}
The navigation subsystem heavily interacts with other spacecraft subsystems, offering the possibility of interconnecting their models for greater situational awareness.
For example, communication and instrument models are used for radiometric and optical navigation, respectively.
Importantly, models residing in different subsystems must be carefully selected to ensure compatibility for cross-subsystem integration and consistency in the type of information they process (see Example~\ref{ex:navigation-uncertainty}).
There is also a need to further explore emerging technologies, such as pulsar-based navigation and optical navigation, and understand how their associated physics models can be integrated to the benefit of system-level autonomy.

\begin{example}{Interconnecting models for greater awareness}{navigation-uncertainty}
    When interconnecting subsystem models, special attention must be paid to their fidelity levels to ensure consistency and continuity of model information across the system~\cite{day2022principles}.
    Consider how \textit{uncertainty} in spacecraft attitude estimates may propagate through a statistical communication model (Section~\ref{sec:comms-high-fidelity}), which in turn may affect radiometric navigation.
    Capturing such effects requires the use of probabilistic models in all three subsystems so that uncertainty can be feasibly propagated and reasoned about.
\end{example}

\section{Communication Subsystem}\label{sec:communications}

The communications subsystem is responsible for receiving and transmitting data between the spacecraft and mission control.
It involves models for antenna gain patterns, signal attenuation, modulation and demodulation processes, and bandwidth management. 
These models are essential for both functional- and system-level autonomy. 
For example, functional-level autonomy might adjust the antenna's orientation and select an optimal modulation scheme to maintain a robust communication link.
System-level autonomy must understand the communication capacity and constraints.
For instance, an autonomous scheduler may need to determine the optimal times to send large data packets or delay non-critical data transmission based on predicted communication link quality and onboard power budgets.
System-level autonomy can also leverage emerging technologies such as demand access~\cite{dhamani2021demand}, which enables spacecraft to request ad-hoc ground support instead of relying on long-lead planning strategies.

The following sections will discuss modeling trades associated with different processes in the communication subsystem.
For a discussion summary, please refer to Table~\ref{tab:communication-subsystem}.

\subsection{Low-fidelity communications modeling}\label{sec:comms-low-fidelity}

\begin{figure}
    \centering
    \includegraphics[width=1.0\linewidth]{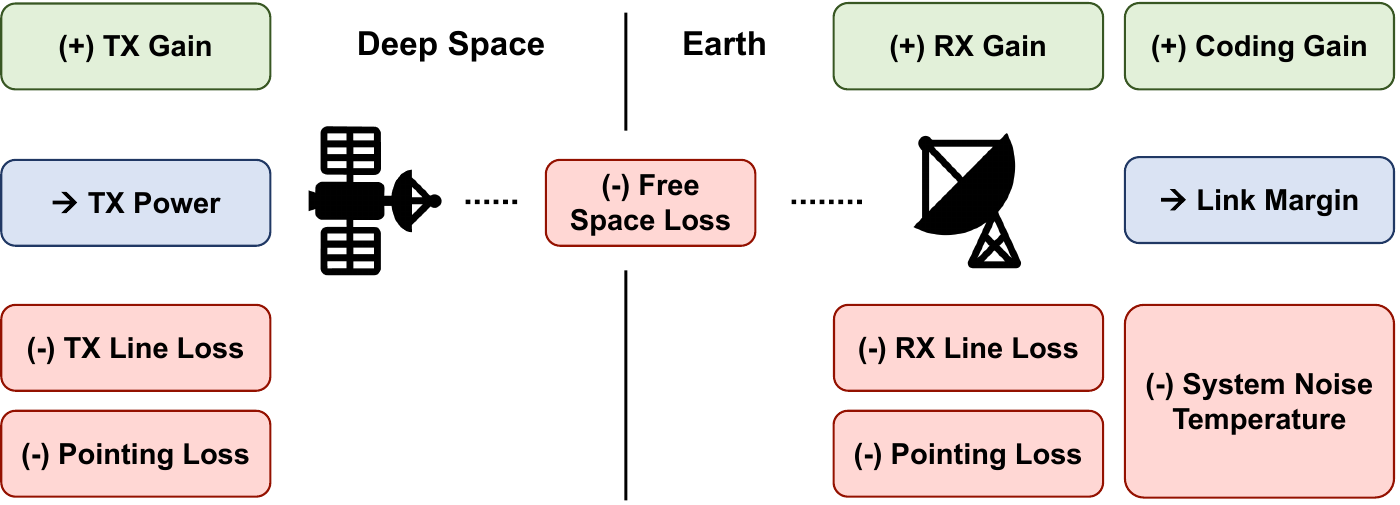}
    \caption{\textbf{Communication subsystem link budget.} The communication link between a satellite and a ground-based antenna is influenced by several parameters, including transmission power, antenna radiation patterns, pointing losses, and atmospheric effects. A link budget analysis considers these deterministic gain and loss parameters along the signal processing chain to compute the receiver carrier-to-noise density ratio, an estimate of the communication link quality.}
    \label{fig:comms-link-budget}
\end{figure}

We begin with a low-fidelity communications model for the communication link quality between a transmitter and receiver.
The model takes a link budgeting approach to compute the received carrier-to-noise density ratio $C/N_0$ ($\si{dB\text{-}Hz}$) by summing deterministic values of link parameters along the signal processing chain (visualized in Figure~\ref{fig:comms-link-budget}):
\begin{equation}\label{eq:carrier-to-noise}
   \frac{C}{N_0} = {\mathrm{EIRP}} + G/T - L - k \quad (\si{dB\text{-}Hz}),
\end{equation}
where $\mathrm{EIRP}$ ($\si{dBW}$) is the equivalent isotropic radiated power, which includes the gains and losses on the transmission side, $G/T$ ($\si{dB\per\kelvin}$) is the receiver gain over system noise temperature, which characterizes the receiver sensitivity, $k$ is the Boltzmann's constant ($-228.6~\si{dBJ\per\kelvin}$), and $L$ ($\si{\decibel}$) represents all losses, including free-space, atmospheric, fading, multi-path, polarization, implementation, and pointing losses.

The data rate, defined as the speed at which data is transmitted over a communication channel in bits per second ($\si{bps}$), plays a key role in communication subsystem design. 
The appropriate data rate is influenced by the acceptable error rate in received data, which can be measured as Bit Error Rate (BER) or Frame Error Rate (FER). 
For example, uncompressed spacecraft telemetry data can tolerate more transmission errors than compressed science data. 
We can infer the necessary energy-per-bit to noise power spectral density ratio ($E_b/N_0$) from coding performance curves by setting an error rate tolerance.
This information, in turn, allows us to model the data rate $R_b$ as:
\begin{equation}\label{eq:data-rate}
    R_b = C/N_0 - E_b/N_0 - M \quad (\si{dB\text{-}HZ}),
\end{equation}
where $R_b\ ($\si{dB\text{-}Hz}$)$ equals $10\log_{10} (R_b \ (\si{bps}))$, $C/N_0$ is the carrier-to-noise density (Eq.~\ref{eq:carrier-to-noise}), and $M$ is the desired link margin. 
Typically, a $\SI{3}{\decibel}$ margin is used to calculate an adequate data rate for a given link design, while larger link margins are used when (a) links have not previously been implemented or (b) if link parameters are not well understood.

While the data rate model given by Eq.~\ref{eq:data-rate} is straightforward and widely accepted for links not constrained by power, we consider it low-fidelity for two reasons.
First, it neglects the statistical nature of specific link parameters, such as polarization and antenna pointing losses~\cite{Cheung2023}.
Instead, the model assumes worst-case values for these link parameters, which may yield overly conservative data rates.
Second, the $\SI{3}{\decibel}$ link margin lacks rigorous mathematical and statistical justification, leaving its adequacy in question for autonomous spacecraft operation in safety-critical settings.

\begin{table}
\renewcommand{\arraystretch}{1.4}
\caption{
    \textbf{Communication subsystem modeling trade.}
    The most prominent tradeoff of low-fidelity communication models is that they do not capture the statistical characteristics of communication links.
    High-fidelity models are also required for higher frequency channels, where non-Guassian and nonlinear effects impact the communication link quality.
}
\label{tab:communication-subsystem}
\centering
\scriptsize
    \begin{tabular}{cll}
    \hline
    \textbf{} & \textbf{Low fidelity communication} & \textbf{High fidelity communication} \\
    \hline
    \hline
    \multirow{5}{*}{\rotatebox[origin=c]{90}{Models}} 
        & Static link budget        & S/C and environment dynamics \\
        & Relevant gains and losses & Non-linear effects and losses \\
        & Deterministic parameters  & Statistical analysis or simulation \\ 
        & Physical OSI layer only   & Data-link and network OSI layers \\
        & Coding and modulation     & Data buffers and protocols \\
    \hline
    \multicolumn{3}{l}{\tiny\bfseries S/C: Spacecraft, OSI: Open Systems Interconnection}
    \end{tabular}
\end{table}

\subsection{Considerations for high-fidelity communications modeling}\label{sec:comms-high-fidelity}

High-fidelity communications modeling focuses on the complexities and statistical nature of communication links~\cite{Cheung2023}.
These models expand on the foundational communication components in low-fidelity models while involving changing carrier-to-noise densities, variations in onboard data generation rates, and onboard data buffering and storage.
Besides improved models of the physical communications layer, high-fidelity modeling of layers above in the OSI model~\cite{isoiec7498} are also core focus areas.
The data link layer, for example, promotes reliable communication through Automatic Repeat reQuest (ARQ) re-transmission strategies~\cite{pollara1995analysis}. 

Let us consider the Go-Back-N ARQ protocol, where  $N$ frames are transmitted before an acknowledgment (ACK) is required from the receiver~\cite{Cheung2023}.
The known acknowledgment channel FER is denoted by $P_{\mathrm{ack}}$, which reflects the probability that acknowledgment messages are received with errors.
We can express the effective data rate $R_{\mathrm{eff}}$ as:
\begin{equation}\label{eq:arq-data-rate}
    R_{\mathrm{e f f}}=\frac{R_b}{1+\frac{N\left(1-\left[1-f\left(\frac{C}{N_0}-R_b\right)\right]\left(1-P_{\mathrm{ack}}\right)\right)}{\left(1-f\left(\frac{C}{N_0}-R_b\right)\right)\left(1-P_{\mathrm{ack}}\right)}} \quad (\si{bps}),
\end{equation}
where $R_b$ is the raw data rate (Eq.~\ref{eq:data-rate}), and $f(\cdot)$ denotes the FER performance curve used in the link. 
Notice how including a data link layer protocol and a known FER may significantly vary the effective communication data rate~\cite{sastry1975improving}.

\begin{example}{Modeling tradeoffs for onboard autonomy}{arq-methods}
    Understanding the assumptions and limitations of low-fidelity models can drive the development of autonomy algorithms that correct for model errors.
    Consider how ARQ onboard autonomy improves communication reliability in the presence of errors that a low-fidelity communication link model does not capture.
    However, the use of ARQ here may also highlight the need to further explore the statistical behavior of communication links.
\end{example}

As mentioned in Example~\ref{ex:navigation-uncertainty}, there are opportunities to refine and interconnect communication subsystem models to capture cross-subsystem effects.
Examples of such effects include how power transmission capacities might influence the communication link quality (Eq.~\ref{eq:carrier-to-noise}) and how uncertainties in attitude estimates might affect antenna pointing losses.
Modeling these dependencies allows for the possibility of onboard situational awareness, whereby system-level autonomy can reason across subsystems to, for example, attribute communication failures to uncertainty in the spacecraft's attitude.
Further understanding these model relationships will help inform future modeling and integration efforts for new or improved autonomous onboard capabilities.

As communications technology progresses to higher frequency channels, such as Ka-band and optical links, the complexities of communication link modeling increase accordingly. 
Such channels are prone to non-Gaussian and nonlinear effects, including disturbances like turbulence, scintillation, and antenna pointing jitter, which may render the constant $C/N_0$ assumption inaccurate for deep space communication. 
For instance, a Ka-band communication link between Earth and a Mars spacecraft, with a round-trip light time of ten to forty minutes and atmospheric disruptions, can experience rapidly varying channels, which cannot be characterized with low-fidelity deterministic models~\cite{Cheung2023}.
Consequently, to ensure the reliability and efficacy of new communications technology for challenging deep space missions, there is an need to further develop high-fidelity communication models.

\section{Discussion}\label{sec:discussion}

\subsection{Key insights: Examples from subsystem fidelity analysis}
\label{sec:discussion-insights}

We have paired our analysis of subsystem model fidelity with inline examples; for instance, Example~\ref{ex:power-solar-array} in the power subsystem.
These examples demonstrate how our modeling considerations can (or have been) applied to benefit their respective subsystems. 
They also provide further insights on modeling and its broader implications on autonomy.
We summarize the key insights of those examples below. 

\begin{itemize}
    \item Example~\ref{ex:power-solar-array}: The need to interconnect models is a byproduct of increasing model fidelity to account for cross-subsystem dependencies and fine-grained environment effects. 
    \item Example~\ref{ex:power-mppt}: Modeling higher-order effects enables the possibility of using autonomy that capitalizes on those effects for greater efficiency or robustness of the system.
    \item Example~\ref{ex:power-battery}: A subsystem process can be described by multiple models, but the autonomy use case(s) of these models may differ according to their implementation.
    \item Example~\ref{ex:attitude-control}: Robustness to higher-order effects can be built into autonomy derived from lower-fidelity models.
    \item Example~\ref{ex:attitude-solar-estimation}: Considering the interconnections among subsystem models can provide opportunities to improve system robustness through cross-checking and direct modeling.
    \item Example~\ref{ex:navigation-uncertainty}: When integrating models across subsystems, choosing models with similar fidelity levels promotes compatibility among models and consistency in the type of information propagated through the system (\textit{e.g.}, uncertainty).
    \item Example~\ref{ex:arq-methods}: The limitations of low-fidelity models can be mitigated via autonomy that corrects for model errors.
\end{itemize}

\subsection{Modeling considerations: Simulation versus onboard}
\label{sec:discussion-modeling-differences}

While simulation software and spacecraft subsystems employ related models, they are unlikely to run identical models in the general case.
This is because model selection is driven by the requirements and constraints of the system, which differ between simulation software and onboard subsystems.
These requirements and constraints can include whether the system (a) must produce or relies on having high-fidelity state predictions, (b) can directly access or indirectly estimate spacecraft and environment states to support model integration, or (c) is constrained by computational complexity, (d) design complexity, and (e) implementation cost.

\subsubsection{Simulation}
The goal of general-purpose spacecraft simulation is to simulate the behavior of the spacecraft, its subsystems, and the environment with high precision and accuracy. 
Here, model selection may not be constrained by computational complexity, affording the use of high-fidelity physics models.
Interconnecting models can be considered a design requirement to support joint simulation of multiple spacecraft subsystems while accounting for cross-subsystem effects.
Importantly, as is the case in simulation, ground-truth access to spacecraft and environment states may simplify the simulator's design and implementation (\textit{i.e.}, model integration).
As noted in Section~\ref{sec:related-work-models}, such a high-fidelity simulation architecture does not currently exist. 
MuSCAT integrates models across several core spacecraft subsystems (Figure~\ref{fig:muscat}), but it specifically targets low-fidelity simulation for efficient prototyping and testing of autonomy algorithms.

\subsubsection{Onboard}
Spacecraft onboard autonomy aims to achieve mission goals independent of external control~\cite{fong2018autonomous}.
Thus, onboard model selection is driven by the enablement of autonomy capabilities required for intended the mission and the satisfaction of mission and spacecraft constraints. 
Mission constraints could, for example, include safety or risk margins, operational efficiency, design, implementation, and operations cost.
Spacecraft constraints include the availability of onboard compute (\textit{i.e.}, computational complexity), hardware, and sensing, along with software design complexity.

The requirements and constraints of a mission may give rise to multiple plausible autonomy architectures (hardware and software), among which the most viable autonomy architecture must be selected.
Figure~\ref{fig:SC-models} depicts one such viable autonomy architecture for the DARE mission (Figure~\ref{fig:SB-mission}), wherein system-level autonomy units plan, manage, and orchestrate the execution of tasks for multiple functional-level autonomy units.
Each autonomy unit at the system-level and functional-level consists of one or more autonomy algorithms running simultaneously.
These autonomy algorithms are carefully selected to meet mission requirements.
Onboard models must be chosen accordingly to support both the design and real-time operation of these autonomy algorithms.
Examples could include the choice of rigid-body dynamics models (Eq.~\ref{eq:att-dynamics-lf}) for attitude control or MPPT models (Eq.~\ref{eq:power-solar-mppt}) offering accurate predictions of generated power for system-level planning.
In general, full spacecraft autonomy will likely require a combination of low-fidelity and high-fidelity models, each supporting their respective autonomy functions.

In contrast to simulation, we cannot assume direct access to spacecraft and environment states, or extensive computational resources onboard the spacecraft.
Such constraints complicate model integration and may bias model selection to efficient (low-fidelity) models.
Nevertheless, the lack of ground-truth state access can, to some extent, be alleviated through state estimation, state prediction, and sensors for direct measurement of states.
Evaluation of the autonomy architecture, its algorithms, and the underlying models in simulation is necessary to quantitatively guide model refinement over multiple testing and development cycles.

\subsection{Connections to Autonomy Principles}
\label{sec:autonomy-principles}
In prior publications, we proposed a set of autonomy principles derived from years of experience architecting autonomous systems~\cite{day2022principles}.
These principles serve to guide the development of future multi-mission autonomous systems.

We discuss the relationship between several autonomy principles and the modeling considerations studied in this work.
Since modeling is our core focus, we only consider the \textit{Information and Knowledge} category of autonomy principles, while principles pertaining to Reasoning, Control Behaviors, and Actions relate to autonomy design.
Please refer to Table~\ref{tab:autonomy-principles} for a description of the relevant autonomy principles.

\begin{table}
\renewcommand{\arraystretch}{1.4}
\caption{
    \textbf{Autonomy principles} relevant to modeling considerations for developing deep space autonomous spacecraft.
}
\label{tab:autonomy-principles}
\centering
\scriptsize
    \begin{tabular}{ll}
    \hline
    \textbf{Principle} & \textbf{Description} \\
    \hline
    \hline
        \begin{tabular}[t]{@{}p{0.36\linewidth}@{}}1.2: Explicit Models\end{tabular} & \begin{tabular}[t]{@{}p{0.54\linewidth}@{}}Use explicit models for information that will be reasoned about\end{tabular} \\ \vspace{0.5pt}
        \begin{tabular}[t]{@{}p{0.36\linewidth}@{}}1.3: Interconnected Models\end{tabular} & \begin{tabular}[t]{@{}p{0.54\linewidth}@{}}Use an interconnected and coordinated set of models across abstractions\end{tabular} \\ \vspace{0.5pt}
        \begin{tabular}[t]{@{}p{0.36\linewidth}@{}}1.4: Abstraction-associated Models\end{tabular} & \begin{tabular}[t]{@{}p{0.54\linewidth}@{}}Associate models with all abstractions that represent data, behaviors, or devices\end{tabular} \\ \vspace{0.5pt}
        \begin{tabular}[t]{@{}p{0.36\linewidth}@{}}1.5: State Access\end{tabular} & \begin{tabular}[t]{@{}p{0.54\linewidth}@{}}Provide a uniform means for identifying, accessing, and managing state knowledge that will be reasoned about\end{tabular} \\ \vspace{0.5pt}
        \begin{tabular}[t]{@{}p{0.36\linewidth}@{}}1.9: State Uncertainty\end{tabular} & \begin{tabular}[t]{@{}p{0.54\linewidth}@{}}Include appropriate characterizations of uncertainty in state knowledge and in associated models\end{tabular} \\
        \begin{tabular}[t]{@{}p{0.36\linewidth}@{}}1.10: Reconciled Knowledge\end{tabular} & \begin{tabular}[t]{@{}p{0.54\linewidth}@{}}Base reasoning on knowledge, where information from considered sources has been reconciled\end{tabular} \\
    \hline
    \end{tabular}
\end{table}

\subsubsection{Principle 1.2: Explicit Models}
\label{sec:principle-explicit-models}
We have outlined explicit models for variables (\textit{e.g.}, states) and processes in four spacecraft subsystems. 
These models describe relevant relationships among subsystem states, enabling (a) functional-level autonomy that estimates, predicts, and controls states and (b) system-level autonomy that reasons over explicit models for system health and data management, planning, and execution. 

\subsubsection{Principle 1.3: Interconnected Models}
\label{sec:principle-interconnected-models}
We have noted several interconnections among subsystem models, particularly where models represent related elements of the same \textit{entity}.
For example, both star trackers and solar array voltage measurements inform the attitude state of the spacecraft. 
Such dependencies offer redundancy on system states (Example~\ref{ex:attitude-solar-estimation}) and the opportunity to reconcile state information (Principle~1.10), both of which support robustness. 
We also outline a hierarchy of fidelity levels to ensure the appropriate use of models for various autonomy functions in the stack.

\subsubsection{Principle 1.4: Abstraction-associated Models}
\label{sec:principle-abstraction-models}
We have associated models with several important abstractions of spacecraft subsystems. 
Associating models with \textit{all} abstractions that will be reasoned about---including data, behaviors, and devices---will require a detailed specification of our spacecraft autonomy architecture. 
This detailed specification does not currently exist but will be outlined in future work.

\subsubsection{Principle 1.5: State Access}
\label{sec:principle-state-access}
As illustrated in Figure~\ref{fig:SC-models}, this principle is honored by our proposed autonomy architecture, which separates system-level autonomy from the underlying subsystems it must operate over and allows system-level autonomy to access the states of those subsystems for reasoning tasks such as system health management and planning.

\subsubsection{Principle 1.9: State Uncertainty}
\label{sec:principle-state-uncertainty}
In our study, accounting for uncertainty in state knowledge and in associated models is exemplified by (a) the use of filtering-based estimators in attitude GNC and navigation and (b) the need to further develop statistical models in communications. 
Propagation of uncertainty across subsystems may be enabled through an interconnected set of models (Example~\ref{ex:navigation-uncertainty}), which informs state and promotes safe decision-making at the system-level.

\subsubsection{Principle 1.10: Reconciled Knowledge}
\label{sec:principle-reconciled-knowledge}
As previously noted, understanding the interconnections among models (Principle~1.3) allows for cross-checking to reconcile knowledge or direct modeling to aggregate knowledge produced by different subsystems. 
Discrepancies in knowledge may reflect faults, while correspondences can be exploited for safe action.

\subsection{Limitations and future work}
We outline four avenues for future work. 
First, our analysis can be extended by implementing higher-fidelity models and evaluating them alongside low-fidelity models.
Such a study could inform onboard model selection based on observed performance gaps and constraints.
Second, while we categorize models based on their fidelity levels, follow-on work should target the subselection of models to satisfy detailed specifications of autonomy requirements and spacecraft constraints.
Third, while we highlight thermal dependencies in several subsystems, we do not analyze thermal models.
The effects of thermals on adjacent subsystems can be significant and should be investigated.
Lastly, our case study considers only the cruise and approach phases of the small-body exploration mission. 
Future work should include downstream mission phases where other subsystems, such as thermal, mobility, and instruments, have a more significant role.

\section{Conclusion}\label{sec:conclusion}

Autonomous spacecraft operation provides an opportunity to expand the frontier of science missions to worlds with complex and dynamic environments that are not well characterized \textit{a priori}.
These missions require functional-level and system-level autonomous capabilities, both of which rely on models to predict subsystem behavior subject to environment effects. 
System-level capabilities, in particular, require situational awareness in order to autonomously plan and schedule tasks, whilst ensuring system health in off-nominal and uncertain scenarios. 
This may drive the need to further refine and interconnect onboard models for greater system-level understanding, in turn enabling robust action.

The focus of this paper is to identify and analyze the tradeoffs of \textit{model fidelity} across four major spacecraft subsystems: power, attitude GNC, navigation, and communications.
Our contributions include the following: (i) an analysis of subsystem models, including categorizations based on model fidelity and couplings with models in adjacent subsystems; (ii) a discussion of modeling considerations to be quantitatively evaluated in future trade studies; (iii) a case study for the cruise-approach phases of a deep space exploration mission, which demonstrates the use of an interconnected set of low-fidelity models for autonomous small-body rendezvous.
In providing a means to inform state knowledge through cross-checking, we conclude that interconnecting models is a promising avenue for developing onboard situational awareness and robust system-level autonomous capabilities.

In summary, this work reflects a first step in the direction of a holistic, system-level assessment of onboard models required for fully autonomous spacecraft operation.
We hope to spur continued research at the intersection of modeling, architecture, and algorithm development for deep space autonomy.

\acknowledgments
The authors would like to thank Martin Cacan and Shyam Bhaskaran for their valuable advice on this paper. This research was carried out at the Jet Propulsion Laboratory, California Institute of Technology, under a contract with the National Aeronautics and Space Administration. 
This work was supported by the National Aeronautics and Space Administration under the Innovative Advanced Concepts (NIAC) program and the ``la Caixa'' Foundation fellowship (ID 100010434, code LCF/BQ/EU21/11890112).

\appendix{}

In this section, we outline MuSCAT's environment models.
These environment models produce disturbance forces and torques on the spacecraft that we refer to in the main text.

\subsection{Gravity disturbance models}\label{appx:gravity}

\subsubsection{Force}\label{appx:gravity-force}
When the spacecraft is positioned at large distances from the Sun and potential small bodies of interest, we simulate the force of gravity as a uniform force field induced by a point mass at the center of a small body $\refm{b}$ (or the Sun $\refm{s}$) acting on the center of mass $\refm{cm}$ of the spacecraft:
\begin{equation}\label{eq:gravity-force-far}
    \tram{F}{b}{cm}{i}^g = -\frac{\mu_b m_{sc}}{||\tram{r}{b}{cm}{i}||^3} \tram{r}{b}{cm}{i},
\end{equation}
where $\mu_b = G m_b$ is the gravitational constant of a small body, $m_{sc}$ is the mass of the spacecraft, and $\tram{r}{b}{cm}{i}$ is the position vector from the small body to the center of mass of the spacecraft.
The Sun and the target small body are the only celestial bodies included in the simulation.

\subsubsection{Torque}\label{appx:gravity-torque}
The gravitational force induces a disturbance torque as a function of the spacecraft's geometry.
We consider the spacecraft as composed of continuous mass elements $dm$ located a $\vecm{r}{{\refm{b}}}{{dm}}{i} = \tram{r}{b}{sc}{i} + \rotm{R}{sc}{i}\vecm{r}{{\refm{sc}}}{{dm}}{sc}$.
The force acting on each mass element is:
\begin{equation*}\label{eq:gravity-force-dm}
    d\vecm{F}{{\refm{b}}}{{dm}}{i}^g = -\frac{\mu_b dm}{||\vecm{r}{{\refm{b}}}{{dm}}{i}||^3} \vecm{r}{{\refm{b}}}{{dm}}{i}.
\end{equation*}
We can compute the total disturbance torque by integrating over the torque contributions of each mass element:
\begin{equation}\label{eq:gravity-torque-dm}
    \prescript{\refm{sc}}{}{\mathbf{T}}^g = \int\displaylimits_{dm \in \refm{sc}} \vecm{r}{{\refm{b}}}{{dm}}{i} \times \left(\rotm{R}{i}{sc} d\vecm{F}{{\refm{b}}}{{dm}}{i}^g \right) dm
\end{equation}

We approximate the total disturbance torque given by Eq.~\ref{eq:gravity-torque-dm} based on the cuboid geometry of a SmallSat spacecraft.
Assuming that the mass of the spacecraft $m_{sc}$ is uniformly distributed, we discretize the spacecraft into $K^3$ cuboids ($K$ partitions along each axis of frame $\refm{sc}$).
Each partition has a mass $m_k = \frac{m_{sc}}{K^3}$ centered at $\vecm{r}{{\refm{b}}}{k}{i} = \tram{r}{b}{sc}{i} + \rotm{R}{sc}{i} \vecm{r}{{\refm{sc}}}{k}{sc}$.
The gravitational force $\vecm{F}{{\refm{b}}}{k}{i}^g$ acting on the $k$-th cuboid and the resulting torque $\prescript{\refm{sc}}{}{\mathbf{T}}^g_k$ is given by:
\begin{gather*}
    \vecm{F}{{\refm{b}}}{k}{i}^g = -\frac{\mu_b m_k}{||\vecm{r}{{\refm{b}}}{k}{i}||^3} \cdot \vecm{r}{{\refm{b}}}{k}{i}, \\
    \prescript{\refm{sc}}{}{\mathbf{T}}^g_k = \vecm{r}{{\refm{cm}}}{k}{sc} \times \left(\rotm{R}{i}{sc} \vecm{F}{{\refm{b}}}{k}{i}^g\right).
\end{gather*}
Finally, the total disturbance torque is computed as the sum of individual disturbance torques of each mass partition:
\begin{equation}\label{eq:gravity-net-torque-approx}
    \prescript{\refm{sc}}{}{\mathbf{T}}^g = \sum_{k=1}^{K^3} \prescript{\refm{sc}}{}{\mathbf{T}}^g_k.
\end{equation}

\subsection{Solar radiation pressure models}\label{appx:solar-radiation-pressure}

\subsubsection{Preliminaries}\label{appx:light-properties}
The energy $E(\lambda)$ produced by light is inversely proportional to its wavelength $\lambda$ (or proportional to its frequency $f$):
\begin{equation*}
    E(\lambda) = \frac{hc}{\lambda} = hf \quad (\si{\joule}),
\end{equation*}
where $h$ is the Planck constant and $c$ is the speed of light.
The flux $\Phi(\lambda)$ in $\si{\per\second\per\square\meter}$ quantifies the number of photons passing through a unit area per second.
Power density $D(\lambda)$ is computed by multiplying flux $\Phi(\lambda)$ with light energy $E(\lambda)$:
\begin{equation*}
    D(\lambda) = \Phi(\lambda) E(\lambda) \quad (\si{\watt\per\square\meter}).
\end{equation*}
The power density (\textit{i.e.}, irradiance) quantifies the energy transferred by the light per second per unit area.
The spectral irradiance $F(\lambda)$ is defined as the wavelength derivative of the power density $\frac{d}{d\lambda}D(\lambda)$.
Because the Sun produces a range of electromagnetic waves with varying wavelengths, we must integrate over the spectrum of wavelengths to compute the total power density of the Sun:
\begin{equation*}
    H_0 = \int_{\forall \lambda} F(\lambda) d\lambda \quad (\si{\watt\per\square\meter}).
\end{equation*}
Here, $H_0$ is a constant that refers to the power density per unit area of the Sun's spherical surface $4\pi r_0^2$, where $r_0$ is the radius of the Sun. 
The power density of sunlight at a distance $d$ from the Sun is computed by taking the ratio of two spherical areas:
\begin{equation}\label{eq:light-irradiance}
    H(d) = H_0 \left(\frac{4\pi r_0^2}{4\pi d^2}\right) = H_0 \left(\frac{r_0}{d}\right)^2 \quad (\si{\watt\per\square\meter}).
\end{equation}
Finally, dividing the power density $H(d)$ by the speed of light $c$ gives the solar radiation pressure (SRP):
\begin{equation}\label{eq:light-srp}
    \rho(d) = \frac{H(d)}{c} \quad (\si{\newton\per\square\meter}).
\end{equation}

\subsubsection{Force}\label{subsec:srp-force}
The solar radiation pressure (Eq.~\ref{eq:light-srp}) exerts a force on the spacecraft as a function of its geometry. 
We assume that the exterior geometry of the spacecraft can be described by a collection of $N$ plates~\cite{JEAN2019167}, each with area $a_n$, reflection coefficient $r_n$, surface normal unit vector $\prescript{\refm{sc}}{}{\mathbf{\hat{n}}}_n$, and center of pressure located at $\vecm{r}{{\refm{s}}}{n}{i} = \tram{r}{s}{sc}{i} + \rotm{R}{sc}{i} \vecm{r}{{\refm{sc}}}{n}{sc}$. 

The SRP force acting on each face is proportional to the incidence angle of the light:
\begin{equation*}\label{eq:srp-angle}
    \theta_n = \arccos\left(\frac{\vecm{r}{{\refm{s}}}{n}{i}}{||\vecm{r}{{\refm{s}}}{n}{i}||^2} \cdot \rotm{R}{sc}{i} \prescript{\refm{sc}}{}{\mathbf{\hat{n}}}_n \right).
\end{equation*}
The SRP force acting on the $n$-th face is given by:
\begin{equation}\label{eq:srp-force}
    \vecm{F}{{\refm{s}}}{n}{i}^\rho = \rho\left(||\tram{r}{s}{sc}{i}||\right) a_n (1 + r_n) \cos(\theta_n) \vecm{r}{{\refm{s}}}{n}{i},
\end{equation}
where $\tram{r}{s}{sc}{i}$ is the position vector from the Sun to the spacecraft. 
Finally, the total SRP force is the sum of individual forces acting on each face:
\begin{equation}\label{eq:srp-net-force}
    \prescript{\refm{i}}{}{\mathbf{F}^\rho} = \sum_{n=1}^N \vecm{F}{{\refm{s}}}{n}{i}^\rho.
\end{equation}
The resulting direction of the total SRP force $\prescript{\refm{i}}{}{\mathbf{F}^\rho}$ depends on the position, attitude, and geometry of the spacecraft.

\subsubsection{Torque}\label{subsec:srp-torque}
The SRP force acting on each face of the spacecraft (Eq.~\ref{eq:srp-force}) induces a disturbance torque about the spacecraft's center of mass.
The torque contributed by the $n$-th face is:
\begin{equation*}\label{eq:srp-torque}
    \prescript{\refm{sc}}{}{\mathbf{T}}^\rho_n = \vecm{r}{{\refm{cm}}}{n}{sc} \times \left(\rotm{R}{i}{sc} \vecm{F}{{\refm{s}}}{n}{i}^\rho \right).
\end{equation*}
The total SRP disturbance torque is finally expressed as the sum of individual torques of each face:
\begin{equation}\label{eq:srp-net-torque}
    \prescript{\refm{sc}}{}{\mathbf{T}}^\rho = \sum_{n=1}^N  \prescript{\refm{sc}}{}{\mathbf{T}}^\rho_n.
\end{equation}

\bibliographystyle{IEEEtran}

\thebiography

\begin{biographywithpic}
{Christopher Agia}{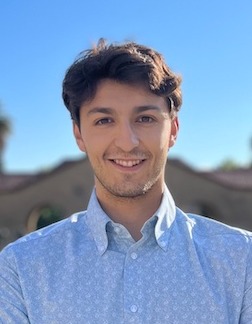} is a visiting researcher at the NASA Jet Propulsion Laboratory and a Ph.D. candidate in the Department of Computer Science at Stanford University, where he works jointly with the Interactive Perception and Robot Learning Laboratory (IPRL) and the Autonomous Systems Laboratory (ASL). He received his B.A.Sc. degree in Engineering Science, Robotics from the University of Toronto in 2021. His research focuses on the development of learning-based methodologies for planning, decision making, and control of robotic manipulation and aerospace systems. He is a member of the Stanford Artificial Intelligence Laboratory (SAIL) and the Center for Research on Foundation Models (CRFM) and has held internships at Microsoft Mixed Reality, Google, and Noah’s Ark Research Laboratory.
\end{biographywithpic} 

\begin{biographywithpic}
{Guillem Casadesus Vila}{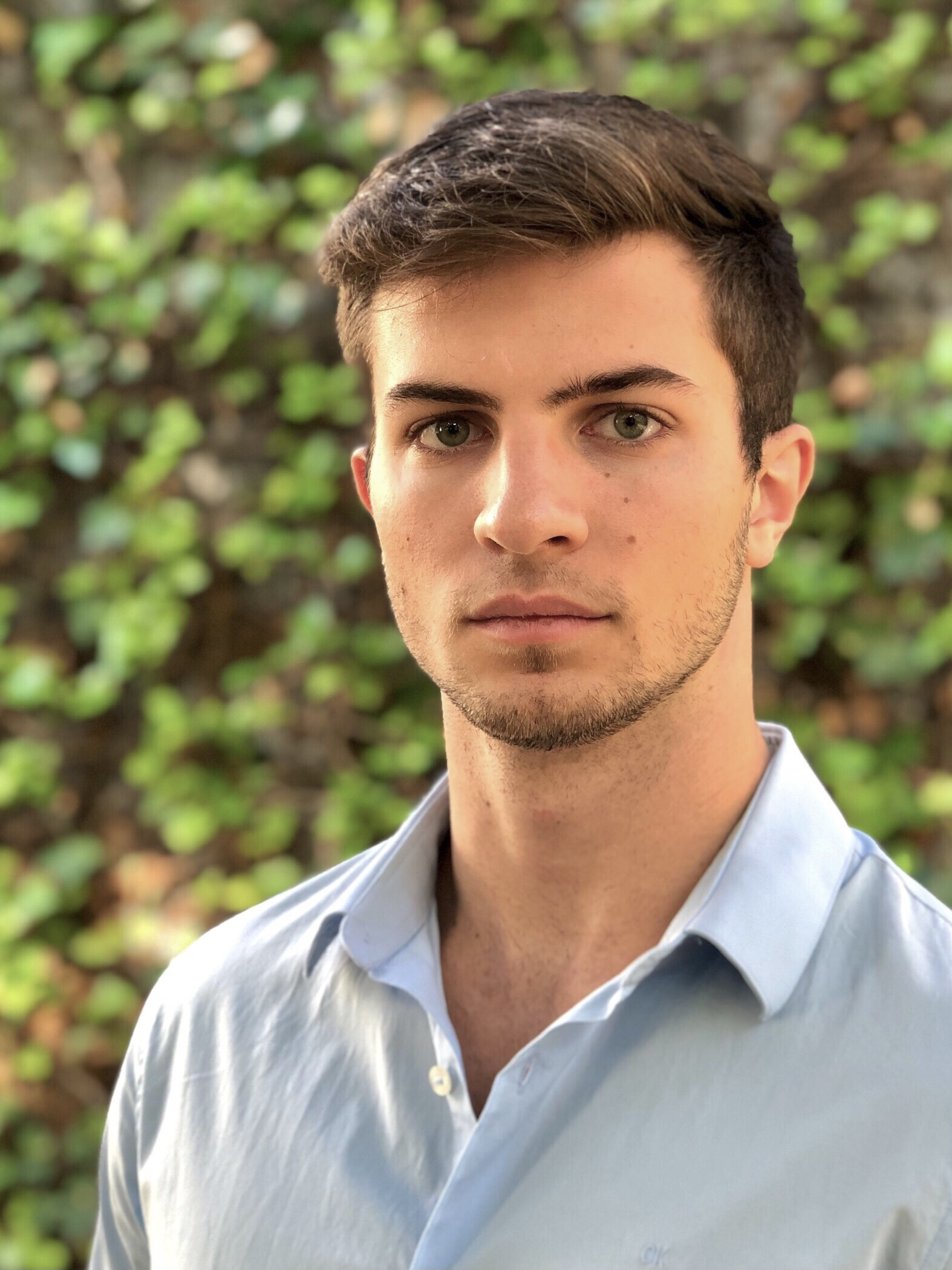} is a Ph.D. student in the Navigation and Autonomous Vehicles (NAV) Lab in the Department of Aeronautics and Astronautics at Stanford University. He received two B.Sc. degrees in Aerospace Engineering and Telecommunications Engineering from the Polytechnic University of Catalonia (UPC) under the Interdisciplinary Higher Education Centre (CFIS) program. Guillem completed his Bachelor’s Thesis at MIT on dynamic resource allocation in multi-beam satellite constellations. He has worked as a researcher at the NASA Jet Propulsion Laboratory, the UPC NanoSat Lab, and the Barcelona Supercomputing Center. His research interests include robotic space exploration, autonomous and distributed space systems, and space communications and navigation.
\end{biographywithpic}

\begin{biographywithpic}
{Saptarshi Bandyopadhyay}{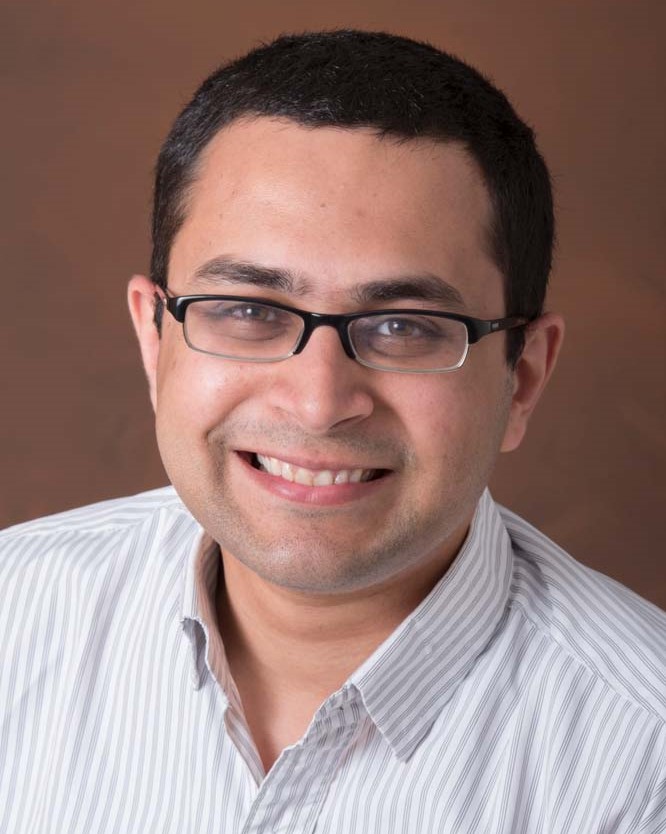} is a Robotics Technologist at the NASA Jet Propulsion Laboratory, where he develops novel algorithms for future multi-agent and swarm missions. He was recently named a NIAC fellow for his work on the Lunar Crater Radio Telescope on the far-side of the Moon. He received his Ph.D. in Aerospace Engineering in 2016 from the University of Illinois at Urbana-Champaign, USA, where he specialized in probabilistic swarm guidance and distributed estimation. He earned his Bachelors and Masters degree in Aerospace Engineering in 2010 from the Indian Institute of Technology Bombay, India, where as an undergraduate, he co-founded and led the institute’s student satellite project Pratham, which was launched into low Earth orbit in September 2016. His engineering expertise stems from a long-standing interest in the science underlying space missions, since winning the gold medal for India at the 9th International Astronomy Olympiad held in Ukraine in 2004. Saptarshi’s current research interests include robotics, multi-agent systems and swarms, dynamics and controls, estimation theory, probability theory, and systems engineering. He has published more than 40 papers in journals and refereed conferences.
\end{biographywithpic}

\begin{biographywithpic}
{David S. Bayard}{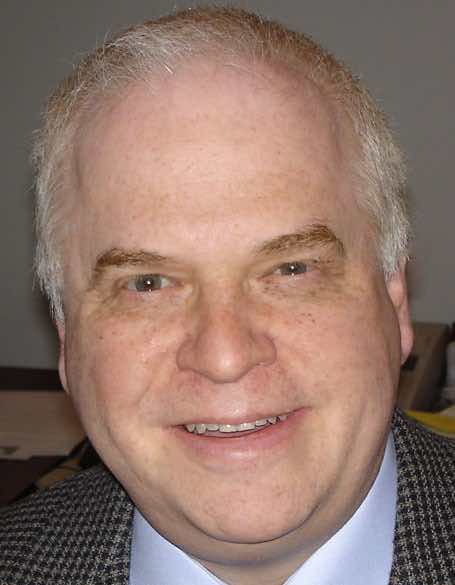} 
is a JPL Senior Research Scientist and Technical Fellow with 38 plus years experience in the aerospace industry. At JPL, he has been involved in the application of modern estimation and control techniques to a range of emerging spacecraft and planetary missions. His work includes over 200 papers in refereed journals and conferences and 4 U.S. patents. He served as an Associate Editor of the IEEE Transactions on Control System Technology (2000-2003), and on the AIAA Technical Committee on Multi-Disciplinary Optimization (1998-2002). Dr. Bayard received numerous NASA awards and achievement medals, and was the recipient of the American Automatic Control Council's (AACC) Control Engineering Practice Award. He is an Associate Fellow of AIAA.
\end{biographywithpic}

\begin{biographywithpic}
{Dr. Kar-Ming Chueng}{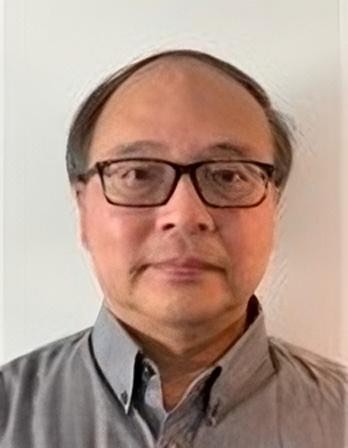} is a Principal Engineer and Technical Group Supervisor in the Communication Architectures and Research Section (332) at JPL. His group supports design and specification of future deep-space and near-Earth communication systems and architectures. He got his B.S.E.E. degree from the University of Michigan, Ann Arbor in 1984, his M.S. degree and Ph.D. degree from California Institute of Technology in 1985 and 1987 respectively.
\end{biographywithpic} 

\begin{biographywithpic}
{Prof. Charles H. Lee}{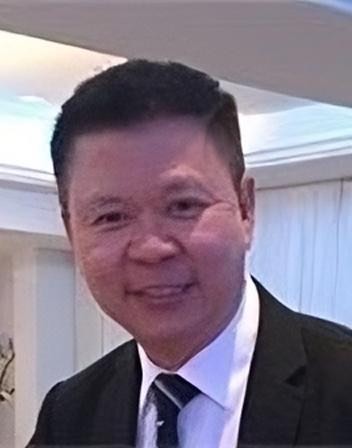}
received his Doctor of Philosophy degree in Applied Mathematics in 1996 from the University of California at Irvine. He then spent three years as a Post-Doctorate Fellow at the Center for Research in Scientific Computation, Raleigh, North Carolina, where he received his 1997-1999 National Science Foundation Industrial Post-Doctorate Fellowship. He became an Assistant Professor of Applied Mathematics at the California State University Fullerton in 1999, Tenured Associate Professor in 2005, and since 2011 he has been a Full Professor. Dr. Lee has collaborated with scientists and engineers at NASA Jet Propulsion Laboratory since 2000. His research has resulted in over 100 professionally refereed articles in Aerospace Engineering, Biomedical Engineering, Bioinformatics, Acoustic Vector Sensors, and Telecommunications. Dr. Lee received Outstanding Paper Awards from the International Congress on Biological and Medical Engineering in 2002 and the International Conference on Computer Graphics and Digital Image Processing in 2017. Dr. Lee received NASA’s Exceptional Public Achievement Medal in 2018 for his Innovative Tools to Assess the Communications \& Architectures Performance of the Mars Relay Network. Dr. Lee also received from JPL an Explorer Award in 2021 and a Voyager Award in 2022 for his innovations in link modeling techniques for collaboration support to Human Landing System and other Lunar and Mars missions and relay network studies.  
\end{biographywithpic} 

\begin{biographywithpic}
{Eric Wood}{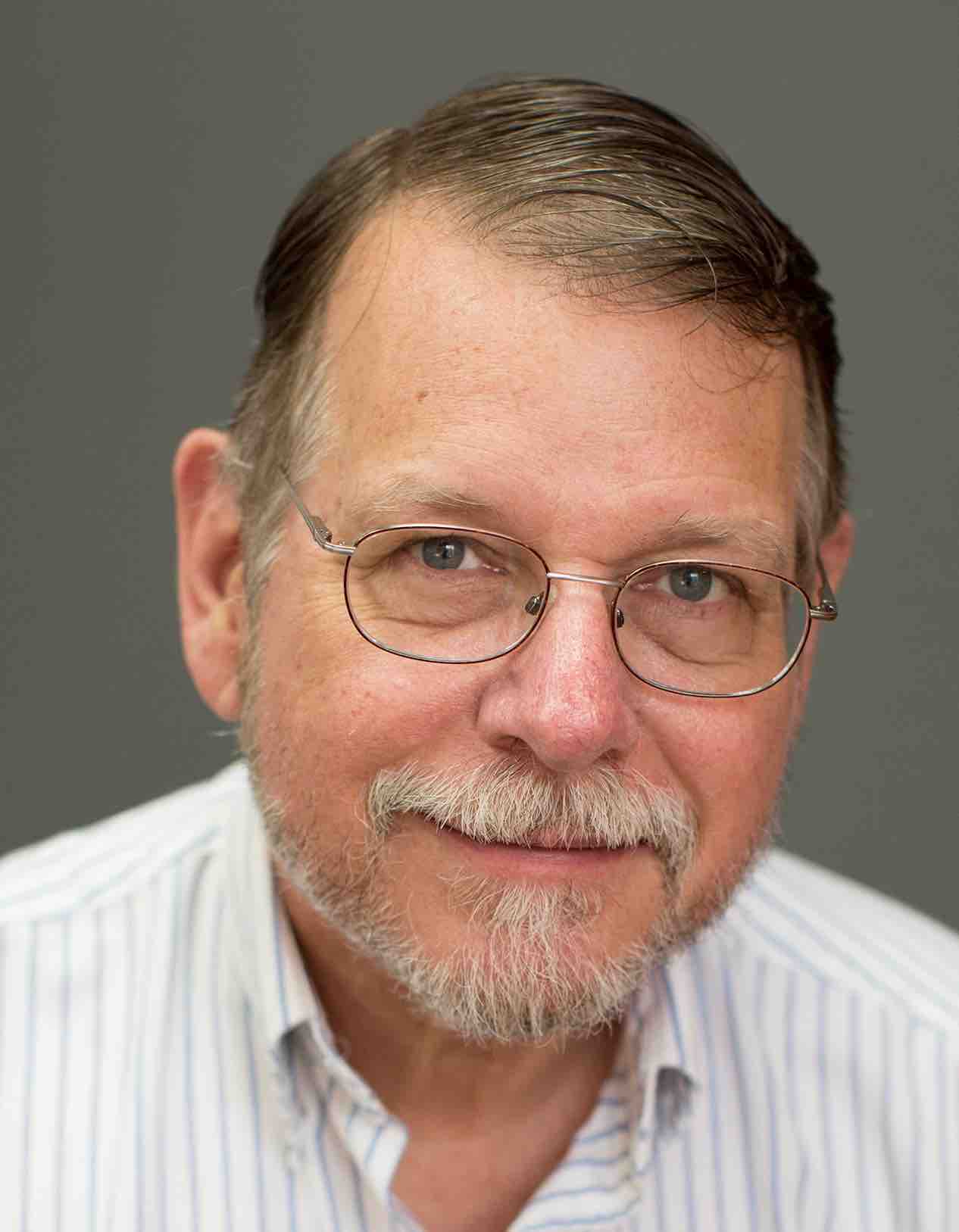} received his B.S. in Computer Science from the California Polytechnic State University at San Luis Obispo in 1974. He is currently the lead developer of the Multi Mission Power Analysis Tool (MMPAT) and the RTG Lifetime Performance Prediction Model (LPPM). He also developed the Spacecraft Power Operations System (SPOS) which is used in ground support of extraterrestrial landed vehicles such as the Spirit, Opportunity, Curiosity and Perseverance Mars rovers.
\end{biographywithpic} 

\begin{biographywithpic}
{Ian Aenishanslin}{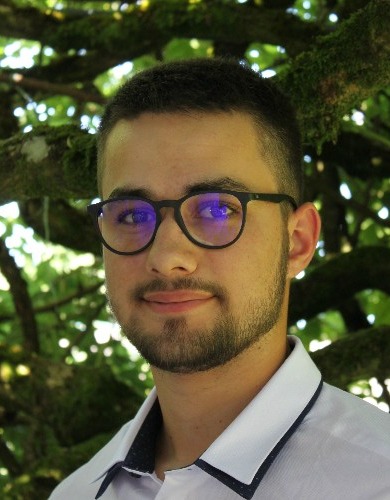} was a space simulation engineer at the NASA Jet Propulsion Laboratory. He received his Bachelor degree in Aerospace Engineering from the Institute of Polytechnic Science and Aeronautics in 2022. During his undergraduate degree, he interned at the National Centre for Space Studies (CNES) where he worked on analog/digital electronic development for processing and control of the instruments onboard the AEROSAT nanosatellite. He also pursued research at the University of Paris where he developed and tested flight software for attitude determination and control of CubeSats. 
\end{biographywithpic}

\begin{biographywithpic}
{Steven Ardito}{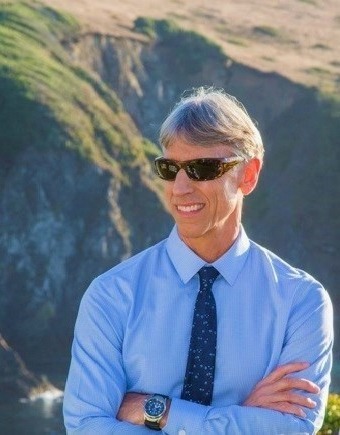} 
is a spacecraft development Chief Engineer with over 30 years experience in system architecture, design, production and operations for Commercial and Government space systems. Extensive technical leadership experience in advanced satellite development, including all system components covering payload, attitude control, propulsion, thermal, power and structure subsystems. Reputation for having uniquely broad and deep skills, allowing for effective management of complex technical development, able to lead diverse teams and achieve pioneering goals. Have also held leadership positions for the development of satellite ground systems, mission operations products, software algorithm design, and missile vehicle design. Excellent track record for successful delivery of highly complex systems, with primary accountability over multiple disciplines including technical risk management and engineering change control.
\end{biographywithpic}

\begin{biographywithpic}
{Lorraine Fesq}{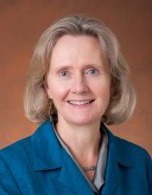} is the Chief Technologist for JPL’s Mission Systems and Operations Division, and the Program Area Manager for Small Bodies and Planetary Defense within JPL’s Planetary Science Directorate. As the recent Program Manager of the ASTERIA CubeSat mission, she demonstrated in-space autonomy experiments. She has over 40 years of experience in industry, government, and academia, working all mission phases from formulation to mission operations. Lorraine holds two patents in autonomy and received a NASA Public Service Medal for her work on the Chandra X-ray Observatory and a NASA Exceptional Achievement Medal for advancing the Fault Management discipline. Lorraine taught at MIT and worked on multiple flight missions at JPL, TRW (now Northrop Grumman), Ball Aerospace and NASA Goddard Space Flight Center. She received her B.A. in Mathematics from Rutgers University and her M.S. and Ph.D. in Computer Science from the University of California, Los Angeles.
\end{biographywithpic}

\begin{biographywithpic}
{Dr. Marco Pavone}{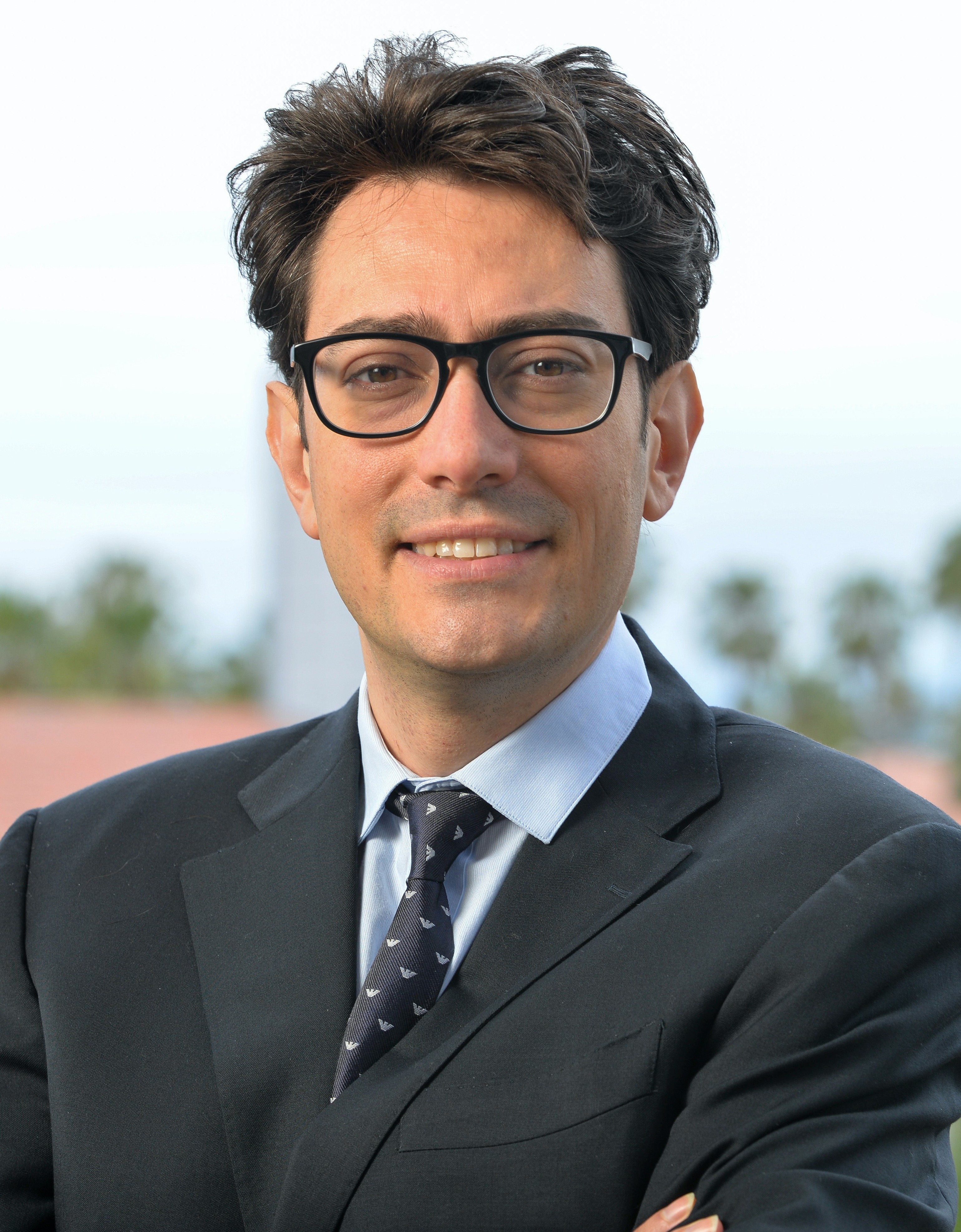} is an Associate Professor of Aeronautics and Astronautics at Stanford University, where he directs the Autonomous Systems Laboratory and the Center for Automotive Research at Stanford. He is also a Distinguished Research Scientist at NVIDIA where he leads autonomous vehicle research. Before joining Stanford, he was a Research Technologist within the Robotics Section at the NASA Jet Propulsion Laboratory. He received a Ph.D. degree in Aeronautics and Astronautics from the Massachusetts Institute of Technology in 2010. His main research interests are in the development of methodologies for the analysis, design, and control of autonomous systems, with an emphasis on self-driving cars, autonomous aerospace vehicles, and future mobility systems. He is a recipient of a number of awards, including a Presidential Early Career Award for Scientists and Engineers from President Barack Obama.
\end{biographywithpic} 

\begin{biographywithpic}
{Issa A.D. Nesnas}{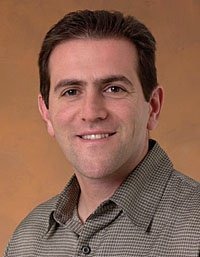} 
is a principal robotics technologist at the Jet Propulsion Laboratory and associate director at Caltech’s Center for Autonomous Systems and Technologies.  He is also the JPL lead on NASA’s Autonomous Systems Capability Leadership Team.  Issa served as the supervisor for the Robotic Mobility Group, which led the development of the autonomous surface navigation for the Perseverance rover. He led research in architecting autonomous systems as well as in-space navigation for approaching small bodies. His other research interests include extreme terrain and microgravity mobility. Issa received a B.E. degree in Electrical Engineering from Manhattan College, NY, in 1991. He earned the M.S. and Ph.D. degrees in Mechanical Engineering from the University of Notre Dame, IN, in 1993 and 1995 respectively. 
\end{biographywithpic}

\end{document}